\documentclass[a4paper,fleqn]{caswenjia}
\usepackage[numbers]{natbib}
\usepackage{geometry}
\usepackage{amsmath}
\usepackage{amsfonts}
\usepackage{amssymb}
\usepackage{bbm}
\usepackage{hyperref}
\usepackage{caption}
\usepackage{graphicx}

\def\tsc#1{\csdef{#1}{\textsc{\lowercase{#1}}\xspace}}
\tsc{WGM}
\tsc{QE}
\tsc{EP}
\tsc{PMS}
\tsc{BEC}
\tsc{DE}

\begin{document}
\begin{sloppypar}

\let\WriteBookmarks\relax
\def\floatpagepagefraction{1}
\def\textpagefraction{.001}

\shorttitle{Physics-Constrained Cross-Resolution Enhancement Network}
\shortauthors{Zhicheng Zhao et~al.}

\title[mode = title]{Physics-Constrained Cross-Resolution Enhancement Network for Optics-Guided Thermal UAV Image Super-Resolution}

\author[1,2]{Zhicheng Zhao}
\author[2,3]{Fengjiao Peng}
\author[2,3]{Jinquan Yan}
\author[3]{Wei Lu}
\author[1,2]{Chenglong Li}
\cormark[1] 
\ead{lcl1314@foxmail.com}
\author[2,3]{Jin Tang}

\address[1]{School of Artificial Intelligence, Anhui University, Hefei, 230601, China}
\address[2]{Anhui Provincial Key Laboratory of Multimodal Cognitive Computation, Anhui University, Hefei 230601, China}
\address[3]{School of Computer Science and Technology, Anhui University, Hefei, 230601, China}

\cortext[cor1]{Corresponding author}

\begin{abstract}
Optics-guided thermal UAV image super-resolution has attracted significant research interest due to its potential in all-weather monitoring applications. However, existing methods typically compress optical features to match thermal feature dimensions for cross-modal alignment and fusion, which not only causes the loss of high-frequency information that is beneficial for thermal super-resolution, but also introduces physically inconsistent artifacts such as texture distortions and edge blurring by overlooking differences in the imaging physics between modalities.
To address these challenges, we propose PCNet to achieve cross-resolution mutual enhancement between optical and thermal modalities, while physically constraining the optical guidance process via thermal conduction to enable robust thermal UAV image super-resolution. 
In particular, we design a Cross-Resolution Mutual Enhancement Module (CRME) to jointly optimize thermal image super-resolution and optical-to-thermal modality conversion, facilitating effective bidirectional feature interaction across resolutions while preserving high-frequency optical priors.
Moreover, we propose a Physics-Driven Thermal Conduction Module (PDTM) that incorporates two-dimensional heat conduction into optical guidance, modeling spatially-varying heat conduction properties to prevent inconsistent artifacts.
In addition, we introduce a temperature consistency loss that enforces regional distribution consistency and boundary gradient smoothness to ensure generated thermal images align with real-world thermal radiation principles. Extensive experiments on VGTSR2.0 and DroneVehicle datasets demonstrate that PCNet significantly outperforms state-of-the-art methods on both reconstruction quality and downstream tasks including semantic segmentation and object detection.
\end{abstract}

\let\printorcid\relax

\maketitle

\section{Introduction}
Unmanned Aerial Vehicle (UAV) thermal imaging generates images by capturing thermal radiation emitted from object surfaces, rendering it robust to environmental variables such as illumination changes and adverse weather conditions~\cite{Hou2022Review}. Consequently, it has emerged as a key technology for all-weather sensing in diverse applications, including military reconnaissance~\cite{Chao2025Navigating}, agricultural monitoring~\cite{Quach2021Real}, power-line inspection~\cite{Anna2024A}, and emergency rescue operations~\cite{Mario2019The}. With their low cost and high mobility, UAVs facilitate rapid, large-scale, and real-time deployment of thermal imaging systems~\cite{Zhao2024CENet}. However, due to the inherent physical limitations of thermal sensors and current technological constraints, images acquired directly by UAVs often suffer from insufficient spatial resolution. This degradation typically manifests as blurred boundaries and a loss of fine textural details~\cite{Vasterling2013Challenges}, severely limiting the interpretive power and practical utility of thermal UAV remote sensing in complex, fine-scale scenarios.

\begin{figure}
    \centering
    \includegraphics[width=.95\linewidth]{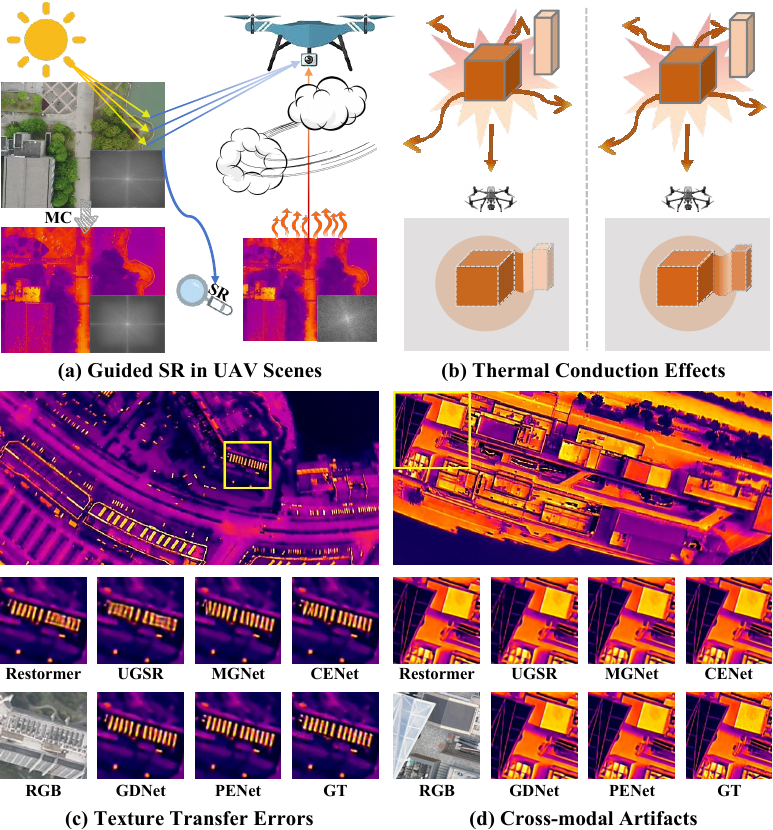}
    \caption{Inaccurate guidance due to the loss of optical high-frequency information and cross-modal imaging discrepancies. We compare SISR methods such as Restormer with GISR methods including UGSR, MGNet, CENet, and GDNet. Our proposed PCNet preserves high-frequency details from optical priors, providing effective guidance for thermal image super-resolution while maintaining high thermophysical consistency between the generated results and the Ground Truth (GT).}
    \label{motivation}
\end{figure}

To address these resolution constraints, researchers have developed various Single Image Super-Resolution (SISR) algorithms~\cite{Dong2014SRCNN,Huang2021Infrared,Dong2022Real-word,LU2025DLMSR,Zhou2023SRFormer}. While such methods enhance image clarity by learning mappings from Low-Resolution (LR) to High-Resolution (HR) domains within the thermal modality, their performance is fundamentally limited by the intrinsic characteristics of thermal images, including low contrast, blurred edges, and lack of textures. These challenges are exacerbated on UAV platforms, where acquired LR thermal images are often degraded by multiple factors such as atmospheric disturbances, sensor noise, and motion blur. Compared to ideal laboratory conditions, the resulting image quality is substantially compromised, which further limits the effectiveness of SISR methods. Consequently, reconstructions frequently lack authentic texture details and fail to preserve physical fidelity. Recognizing this information deficit in thermal images, modern UAV remote sensing platforms are commonly equipped with dual-modality cameras that concurrently capture HR optical images and LR thermal images. Given that optical images contain rich texture and clear edge details absent in the thermal domain, recent research has increasingly focused on exploiting these complementary optical cues to enhance the spatial resolution and perceptual quality of thermal images. As illustrated in Fig.~\ref{motivation}(a), two primary approaches leverage optical images for HR thermal image generation: Guided Image Super-Resolution (GISR)~\cite{Gupta2020Pyramidal,Gupta2021UGSR,Rivadeneira2023Thermal,Arnold2024SwinFuSR,ZHAO2025GDNet}, which employs HR optical images to guide the super-resolution of LR thermal images, and Modality Conversion (MC)~\cite{Isola2017pix2pixGAN,Kniaz2018ThermalGAN,Ordun2022TFC-GAN,Mehmet2022InfraGAN,Lee2023EGGAN}, which learns an end-to-end mapping from HR optical images to HR thermal images.

However, existing methods fail to sufficiently utilize high-frequency optical information and cannot maintain thermophysical consistency. Specifically, to align optical and thermal features, existing approaches typically perform spatial compression on optical features before feeding them into the network at the same resolution as thermal features. This inevitably sacrifices high-frequency information inherent to optical features such as fine textures and clear edges that are beneficial for thermal image super-resolution.
Moreover, these methods often overlook the imaging physics disparities between modalities, causing optical priors to introduce physically inconsistent guidance. As shown in Fig.~\ref{motivation}(a) and (b), optical images primarily capture surface reflections and geometric textures, exhibiting rich high-frequency information, whereas thermal images are governed by thermal radiation and heat conduction, resulting in spatial spectra dominated by lower frequencies and smoothed boundaries. Consequently, the loss of optical high-frequency information and the lack of thermophysical constraints lead to inaccurate guidance, as illustrated in Fig.~\ref{motivation}(c) and (d). Existing methods predominantly exploit optical features through simplified modality alignment and direct texture transfer, which not only potentially discard priors essential for recovering thermal boundaries but also erroneously map optical textures onto the thermal domain. This manifests as stripe distortions on solar panels where thermal distribution should be uniform (Fig.~\ref{motivation}(c)) and discontinuous thermal radiation variations on glass surfaces with complex reflection characteristics (Fig.~\ref{motivation}(d)). Therefore, developing a physics-constrained guided thermal UAV image super-resolution technique that maintains thermal consistency while effectively exploiting optical priors remains a significant challenge.

In this paper, we propose a novel Physics-Constrained Cross-Resolution Enhancement Network (PCNet) for robust guided thermal UAV image super-resolution. PCNet addresses both the high-frequency information loss caused by resolution compression and the physical inconsistency in existing cross-modal guidance by enabling cross-resolution mutual enhancement while constraining optical priors through thermal physics, thereby achieving thermally consistent super-resolution enhancement.
To the best of our knowledge, this is the first physics-constrained cross-resolution network for optics-guided thermal image super-resolution.
To address the loss of high-frequency optical information during the guiding process, we propose a Cross-Resolution Mutual Enhancement Module (CRME), which facilitates comprehensive and efficient cross-resolution information exchange between optical and thermal modalities. This module achieves an efficient super-resolution process by jointly optimizing SR and MC tasks, enabling the network to utilize HR optical priors to guide LR thermal images. Specifically, feature alignment is first performed on different task branches via learnable scale-adaptive projection layers. Subsequently, cross-resolution attention mechanisms fuse cross-modal complementary information, mitigating the loss of optical high-frequency information while sufficiently leveraging the optical prior to guide thermal image super-resolution.
Furthermore, to maintain thermal consistency, we propose a Physics-Driven Thermal Conduction Module (PDTM) to impose physical constraints on the interaction information between modalities in CRME, which embeds two-dimensional heat conduction principles into the optical guidance process, preventing the generation of physically implausible artifacts in the HR thermal images. Additionally, we introduce a Temperature Consistency Loss (TCLoss) that enforces the physical consistency of the temperature distribution through two complementary constraints: regional distribution consistency using Wasserstein distance and boundary smoothness constraints via gradient magnitude regularization. 
Comprehensive experiments on the VGTSR2.0 and DroneVehicle datasets validate the significant superiority of the proposed method. In summary, the main contributions of this paper are as follows:

\begin{itemize}
\item We propose PCNet, the first physics-constrained cross-resolution framework for optics-guided thermal UAV image SR, which enables cross-resolution mutual enhancement while constraining the optics-guided SR process through laws of thermal physics. This design addresses the challenges of optical information loss and thermophysical inconsistencies during the guiding process.

\item We propose the Cross-Resolution Mutual Enhancement Module (CRME) to achieve effective bidirectional feature interaction and information transfer between optical and thermal images at different resolutions, jointly optimizing thermal SR and optical-to-thermal MC to mitigate the loss of high-frequency information in the optical prior during the guidance process.

\item To achieve physical consistency in thermal images, we propose the Physics-Driven Thermal Conduction Module (PDTM), which imposes physical constraints on optical guidance information by embedding a two-dimensional temperature field. Additionally, we introduce a Temperature Consistency Loss (TCLoss) designed to maintain physically plausible temperature distributions.

\item Extensive experiments on the VGTSR2.0 and DroneVehicle datasets demonstrate that PCNet significantly outperforms existing state-of-the-art SISR and GISR approaches in terms of reconstruction quality and downstream tasks, including semantic segmentation and object detection.
\end{itemize}

\section{Related Work}
\label{sec:Related}
This section reviews the studies most relevant to our research, including single image SR, guided image SR, and cross-resolution guided image fusion methods.

\subsection{Single Image SR Methods}
Since SRCNN~\cite{Dong2014SRCNN} pioneered the application of Convolutional Neural Networks (CNN) to super-resolution (SR) tasks, deep learning~\cite{Deng2025decoupled,Deng2025learning,Deng2024collaborative} has become the dominant paradigm in Single Image Super-Resolution (SISR)~\cite{Kim2016VDSR,Lim2017EDSR}. While CNNs excel at local feature extraction, their limited receptive fields constrain long-range dependency modeling. The introduction of attention mechanisms has partially addressed this limitation. RCAN~\cite{Zhang2018RCAN} first integrated channel attention within SR networks, enabling adaptive feature weight adjustment across channels. Subsequent works, such as SAN~\cite{Dai2019SAN} and HAN~\cite{Niu2020HAN}, extended this concept through second-order and hybrid attention mechanisms, respectively, enhancing global feature extraction capabilities. 
The advent of Transformers has further advanced the field. IPT~\cite{Chen2021IPT} first applied pure Transformer architectures to image processing tasks, validating the effectiveness of self-attention mechanisms for image SR. SwinIR~\cite{Liang2021SwinIR} advanced this direction by designing specialized image restoration networks based on Swin Transformers, achieving significant performance improvements through hierarchical windowed self-attention. Restormer~\cite{Zamir2022Restormer} further optimized computational efficiency while maintaining global modeling capabilities via gated feedforward networks and depth-wise convolution attention. HAT~\cite{Chen2023HAT} proposed hybrid attention Transformers integrating channel and self-attention mechanisms to activate broader input pixel ranges for comprehensive global dependency capture. In the context of remote sensing, SPT~\cite{Hao2024SPT} introduced multi-scale perceptual back-projection Transformers with feedback mechanisms to handle scale variations, while CCST~\cite{Yang2025CCST} integrated cross-coding into self-attention to construct long-range dependencies efficiently.

Recent advances have also explored novel generative paradigms. Diffusion models~\cite{Jonathan2020DDPM}, exemplified by SR3~\cite{Chitwan2021SR3}, leverage iterative denoising for high-quality conditional generation. LDM~\cite{Rombach2022LDM} improved efficiency via latent space diffusion, while SeeSR~\cite{Wu2024SeeSR} incorporated semantic priors for enhanced realism. State space models have also gained prominence.
MambaIR~\cite{Guo2024MambaIR} introduced selective state space models for capturing long-range dependencies with linear complexity, and MFEM~\cite{Chen2025MFEM} enhanced Mamba architectures through multi-scale frequency mechanisms to mitigate local information loss.
Specific to thermal SR tasks, TherISuRNet~\cite{Chudasama2020TherISuRNet} employed asymmetric residual learning for frequency-specific feature extraction. DASR~\cite{LIANG2023DASR} utilized Transformers with channel-spatial dual attention for edge structure preservation, while CIPPSRNet~\cite{Wang2022CIPPSRNet} explicitly modeled camera intrinsic parameters. DifIISR~\cite{Li2025DifIISR} advanced the field via gradient-guided diffusion models tailored to infrared spectral characteristics. 

Despite these advances, SISR methods continue to face challenges in generating high-quality HR images, particularly regarding edge preservation and distortion artifacts, challenges that are exacerbated in UAV thermal imaging due to complex operational environments and inherent characteristics such as low thermal contrast.
 
\subsection{Guided Image SR Methods}
To overcome the inherent limitations of SISR methods, researchers proposed Guided Image Super-Resolution (GISR) as a solution. This approach traces back to the joint bilateral upsampling introduced by Kopf et al.~\cite{Kopf2007JBU}, which pioneered the concept of leveraging high-resolution guide images to enhance the quality of low-resolution images. Hui et al.~\cite{Hui2016DMSR} pioneered the application of deep learning to guided image SR tasks. Subsequently, Zhou et al.~\cite{Zhou2017CDSR} further proposed deep guided networks specifically designed to tackle SR challenges in depth images.

In the domain of guided thermal image SR, optical images are predominantly selected as guidance since they possess rich high-frequency information absent in thermal images, such as detailed textures and clear object edges. Chen et al.~\cite{Chen2016LBSR} employed guided filtering~\cite{He2013GIF} to utilize optical images for enhancing the resolution of thermal images. Han et al.~\cite{Han2017CNN-OGT} proposed a CNN-based algorithm that extracts high-frequency information from optical images to supplement the missing texture features in thermal imaging. Pan et al.~\cite{Pan2019Spatially} assumed that the target high-resolution image could be linearly represented by the input low-resolution and guided images, using a CNN network to predict linear coefficients for image reconstruction. Almasri et al.~\cite{Almasri2018GAN-SR} employed a GAN-based model to enhance the resolution of thermal images guided by optical images. CoReFusion~\cite{Kasliwal2023CoReFusion}, built upon UNet, introduced two independent ResNet34 encoders to achieve computationally efficient guided thermal image SR. However, these methods are constrained by limited receptive fields, making it difficult to model long-range spatial dependencies within images.
Given the efficacy of Vision Transformers~\cite{Dosovitskiy2020AnII} in long-range modeling, recent works have adopted Transformer-based GISR approaches. MGNet~\cite{Zhao2023MGNet} extracts diverse feature cues such as appearance, edges and semantics from visible images to guide thermal SR. SwinFuSR~\cite{Arnold2024SwinFuSR} utilizes dual-branch architectures and attention-guided cross-domain fusion. Building on this, SwinPaste~\cite{Zhong2025SwinPaste} enhances generalization through data mixing strategies. 

Despite these developments, a common limitation persists: to enable cross-modal feature alignment and fusion, existing methods typically compress optical features to match thermal feature dimensions, inevitably losing high-frequency information beneficial for thermal image SR. Furthermore, current approaches frequently result in physical distortion in complex scenes due to the neglect of imaging physics differences.

\begin{figure*}
    \centering
    \includegraphics[width=1\linewidth]{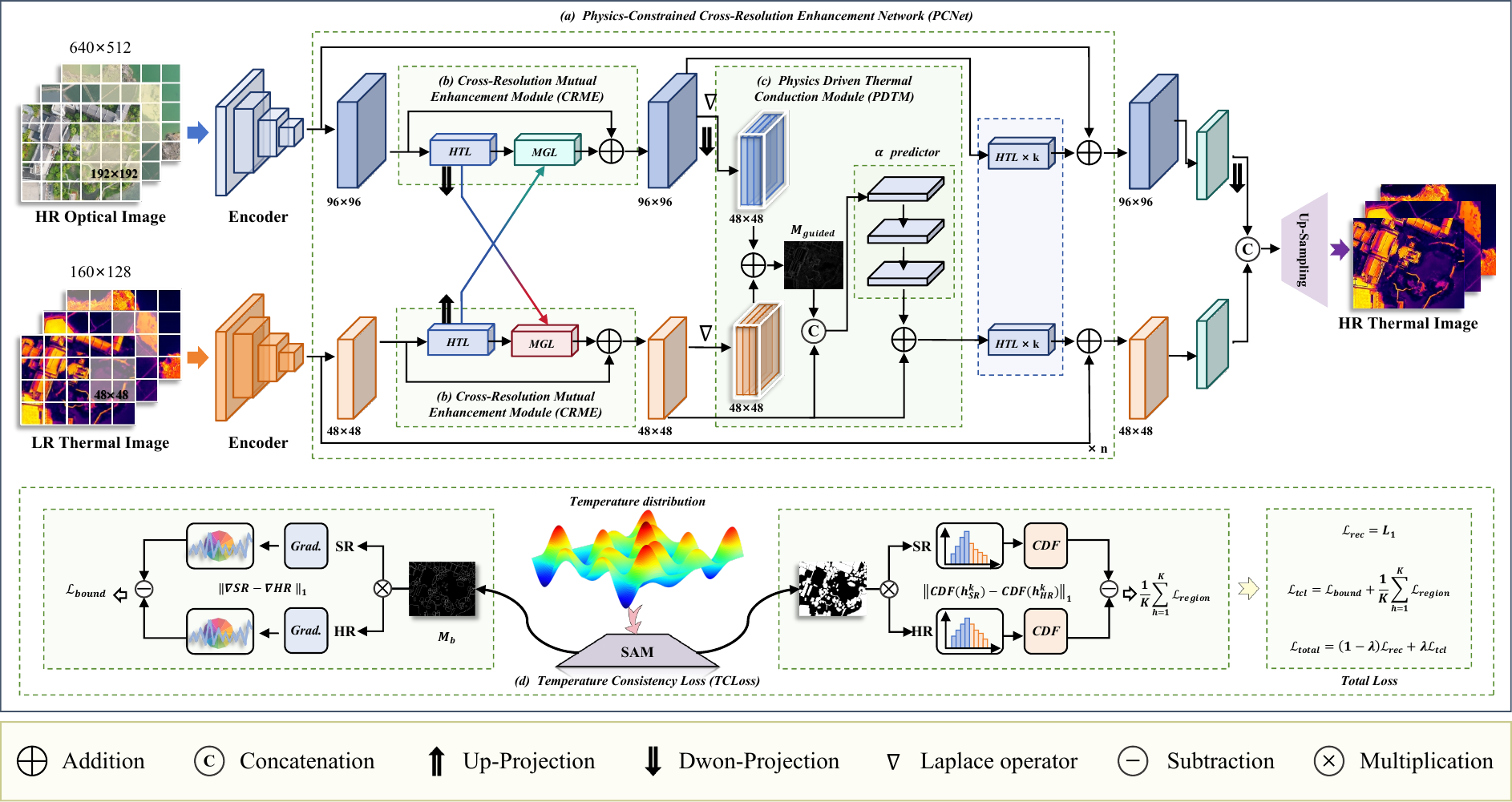}
    \caption{The overall structure of the proposed PCNet. The Cross-Resolution Mutual Enhancement Module (CRME) employs cross-resolution cross-attention to achieve mutual enhancement between modalities while preserving high-frequency optical priors and jointly optimizing SR and MC tasks. The Physics-Driven Thermal Conduction Module (PDTM) achieves physically consistent optical guidance and thermal information enhancement by simulating how thermal images respond to heat diffusion at the edges of real-world objects. Additionally, we incorporate a physical consistency constraint into the loss function to further reduce the discrepancy between SR thermal images and HR thermal images.} 
    \label{overall}
\end{figure*}

\subsection{Cross-Resolution Guided Image Fusion}
Cross-resolution guided image fusion aims to recover spatial details in low-resolution (LR) target images by leveraging structural priors from high-resolution (HR) reference images. The key challenge lies in effectively bridging the resolution gap, transferring high-frequency information such as edges and textures from the HR domain to the LR domain without introducing artifacts or spectral distortion. It has been primarily applied to the reference-based super-resolution (RefSR) task~\cite{Shim2020RefSR,Yang2020RefSR-Transformer,Sun2025DEMSR}, the Cross-resolution person re-identification (CR-ReID)~\cite{Liu2024ITME-ReID,Ouyang2025CR-ReID,Pang2025Robust-ReID} task, the panchromatic sharpening task~\cite{Andrea2007filterPan,Huang2015DNNPan,Zheng2025diffusionPan} and the GISR task~\cite{LU2025medicalSR,Huang2026PCFFusion}.

Early approaches primarily employed naive fusion strategies, such as channel-wise concatenation or element-wise summation, to merge features extracted from parallel branches~\cite{Li2016DJF,Hu2019ACNet}. While computationally efficient, these methods treat HR and LR images equally, failing to address inherent spatial misalignment and semantic discrepancies arising from resolution gaps. To mitigate spatial mismatch, CrossNet++~\cite{Tan2021CrossNet++} combines optical flow estimation with deformable convolutions to rigorously achieve geometric alignment between images. Wang et al.~\cite{Wang2018sftgan} proposed Spatial Feature Transformation (SFT), which generates affine transformation parameters from semantic priors to modulate intermediate features in the recovery network. In remote sensing, Wang et al.~\cite{Wang2025CD} introduced RDENet, employing a dual spatio-temporal resolution-difference modulation (BRDM) strategy to implicitly encode scale variance, while Hong et al.~\cite{Hong2026alignment} explored dual-domain representation alignment to minimize domain bias from resolution differences. Recent research has shifted focus from unidirectional guidance toward dynamic interactions and mutual learning. Li et al.~\cite{Li2025DCILNet} proposed the DCILNet, which achieves demand-space correction between resolutions via cross-resolution feature correction modules. FreeFusion~\cite{Zhao2025FreeFusion} compels the model to capture domain-adaptive representations by reconstructing complementary modalities.

Despite these advances, existing methods employing dual-branch architectures with identical resolutions remain ineffective at bridging the resolution and physical imaging disparities between HR optical and LR thermal images. This not only prevents the full utilization of optical priors but also risks introducing information inconsistent with thermophysical coherence during the guidance process. Consequently, developing a guidance image fusion architecture that enables cross-resolution feature mutual learning under physical constraints is pivotal to solving cross-resolution guided SR.

\section{Methodology}
This section details the proposed Physics-Constrained Cross-Resolution Enhancement Network. Section~\ref{overallframework} presents the overall framework, followed by detailed descriptions of the key internal modules in Sections~\ref{crme} and~\ref{pdtm}. Finally, Section~\ref{loss} discusses the loss function employed in PCNet.

\subsection{Overall Framework}
\label{overallframework}
The architecture of the proposed PCNet is illustrated in Fig.~\ref{overall}, which is a dual-branch physics-constrained cross-resolution enhancement network to generate physically consistent HR thermal UAV images by jointly optimizing thermal SR and optical-to-thermal MC tasks. Specifically, degraded LR thermal and well-aligned HR optical images are randomly sampled and cropped into fixed patches before being simultaneously fed into PCNet for joint training. For thermal images, we employ an encoder with a single $3 \times 3$ convolution layer to extract shallow features and perform channel expansion:
\begin{equation}
F_{T_{initial}} = \operatorname{Conv}(I_{thermal}) \in \mathbb{R}^{H \times W \times C},
\end{equation}
where $I_{thermal} \in \mathbb{R}^{H \times W \times 3}$ denotes the input LR thermal image, and $\operatorname{Conv}(\cdot)$ represents the $3 \times 3$ convolution operation. Given the higher resolution of optical images, we utilize an encoder with three consecutive convolution layers to extract rich high-frequency information of optical images: 
\begin{equation}
F_{O_{initial}} = \operatorname{Conv}(\operatorname{Conv}^{\downarrow}(\operatorname{Conv}(I_{optical}))) \in \mathbb{R}^{2H \times 2W \times C},
\end{equation}
where $\operatorname{Conv}^{\downarrow}(\cdot)$ indicates stride-2 convolution. To balance network computing efficiency with the preservation of more optical high-frequency information, $F_{O_{initial}}$ maintains doubled spatial resolution relative to thermal features $F_{T_{initial}}$, retaining high-frequency details essential for cross-modal guidance as much as possible. 
Following shallow feature extraction, we employ $N$ cascaded enhancement stages, each consisting of two branches. In each stage, the shallow optical and thermal features are first fed into the Cross-Resolution Mutual Enhancement (CRME) module, which promotes bidirectional guidance between modalities. The updated features are then jointly processed by a Physics-Driven Thermal Conduction Module (PDTM) to enforce physically consistent optical guidance on the thermal branch, and finally each branch passes through $K$ Hierarchical Transformer Layers (HTL) to extract deeper feature representations. For the $i$-th stage ($i = 1, 2, \ldots, N$), the processing pipeline proceeds as follows: Optical and thermal shallow features are fed into the CRME for cross-resolution and cross-modal mutual feature enhancement, where optical features provide high-frequency guidance for thermal SR, while thermal features offer thermal information for optical-to-thermal MC:
\begin{equation}
\begin{aligned}
\tilde{F}_{T,i} &= f_{CRME_{SR},i}(F_{T_{i-1}},F_{O_{i-1}}),\\
\tilde{F}_{O,i} &= f_{CRME_{MC},i}(F_{O_{i-1}}, F_{T_{i-1}}),
\end{aligned}
\end{equation}
where $f_{CRME_{SR},i}(\cdot)$ and $f_{CRME_{MC},i}(\cdot)$ denote the $i$-th CRME on the SR branch and the MC branch. $\tilde{F}_{T,i}$ and $\tilde{F}_{O,i}$ are the mutually interacted thermal and optical features. Subsequently, to ensure that the transferred optical details conform to thermal physics, the bidirectionally interacted thermal and optical features are fed into the PDTM, which enhances thermal features along real object boundaries by leveraging optical edge information as geometric constraints:
\begin{equation}
F_{Te,i} = f_{PDTM,i}(\tilde{F}_{T,i}, \nabla^2 \tilde{F}_{O,i}) + \tilde{F}_{T,i},
\end{equation}
where $\nabla^2 \tilde{F}_{O,i}$ represents second-order differential edge features that encode thermal conduction boundaries and $f_{PDTM,i}(\cdot)$ denotes the $i$-th PDTM. Finally, the CRME-enhanced optical features $\tilde{F}_{O,i}$ and PDTM-Constrained thermal features $F_{Te,i}$ are fed into multiple cascaded HTLs, which allows them to extract deeper feature maps and concentrate more on their respective tasks:
\begin{equation}
\begin{aligned}
F_{T_{i}} &= f_{HTL_{K},i}(F_{Te,i}) + F_{T_{i-1}},\\
F_{O_{i}} &= f_{HTL_{K},i}(\tilde{F}_{O,i}) + F_{O_{i-1}},
\end{aligned}
\end{equation}
where $f_{HTL_{K},i}(\cdot)$ denotes a stack of $K$ HTLs in stage $i$. Recognizing the complementary nature of super-resolution and modality conversion tasks, we concatenate the optical and thermal features processed by the $N$ stages as the output features for the SR branch, then perform upsampling to generate the final SR results as follows:
\begin{align}
F_{f} &= \text{Conv}(\text{Concat}(F_{T},F_{O})), \\
I_{SR} &= \text{Conv}(f_{pixel-shuffle}(\text{Conv}(F_{f}))),
\end{align}
where $F_{T}$ and $F_{O}$ are the output thermal feature and optical feature of the last stage on the SR branch and the MC branch, $\text{Concat}(\cdot)$ represents concatenating two features along the channel dimension, $F_{f}$ is the Fused Feature, and $f_{pixel-shuffle}(\cdot)$ denotes Pixel-Shuffle upsampling method.

\subsection{Cross-Resolution Mutual Enhancement Module}
\label{crme}
Existing multi-modal super-resolution methods often rely on single-stream fusion or dual-stream architectures that operate at the same low resolution (LR). These approaches typically downsample high-resolution (HR) optical images to match the thermal input, resulting in the loss of critical high-frequency details like edges and textures at the network input stage. To overcome this, PCNet employs a dual-branch structure that maintains distinct resolutions for the Super-Resolution (SR) and Modality Conversion (MC) tasks. The CRME module is designed to bridge these branches, facilitating joint optimization and bidirectional information flow, as shown in Fig.~\ref{overall}(b).
Crucially, the MC branch processes optical features at a higher resolution (HR), preserving the rich high-frequency information inherent in the optical modality. The CRME module facilitates interaction between the two branches by projecting the preserved HR optical priors into the thermal feature space. This mechanism injects precise high-frequency cues into the SR branch for detail reconstruction while leveraging thermal semantic information to align optical features with the thermal distribution. By enabling this bidirectional cross-resolution interaction, PCNet effectively mitigates the information loss associated with iso-resolution fusion and progressively aligns thermal features with their high-resolution counterparts.
As illustrated in Fig.~\ref{crme}, On each branch, CRME first employs a HTL to capture long-range dependencies and global context. This is followed by a learnable Scale-Adaptive Projection Layer (SPL) for dynamic feature alignment and a Mutual Guidance Attention Layer (MGL) to achieve mutual enhancement:
\begin{equation}
\begin{aligned}
\tilde{F}_{T,i} &= f_{MGL}(f_{HTL}(F_{T_{i-1}}),f_{SPT_{SR}}(f_{HTL}(F_{O_{i-1}}))),\\
\tilde{F}_{O,i} &= f_{MGL}(f_{HTL}(F_{O_{i-1}}),f_{SPT_{MC}}(f_{HTL}(F_{T_{i-1}}))),
\end{aligned}
\end{equation}
where $f_{SPT_{SR}}(\cdot)$ and $f_{SPT_{MC}}(\cdot)$ represent different SPL operations on the SR branch and MC branch respectively. The specific operations for each component are as follows.

\begin{figure}[t]
    \centering
    \includegraphics[width=0.85\linewidth]{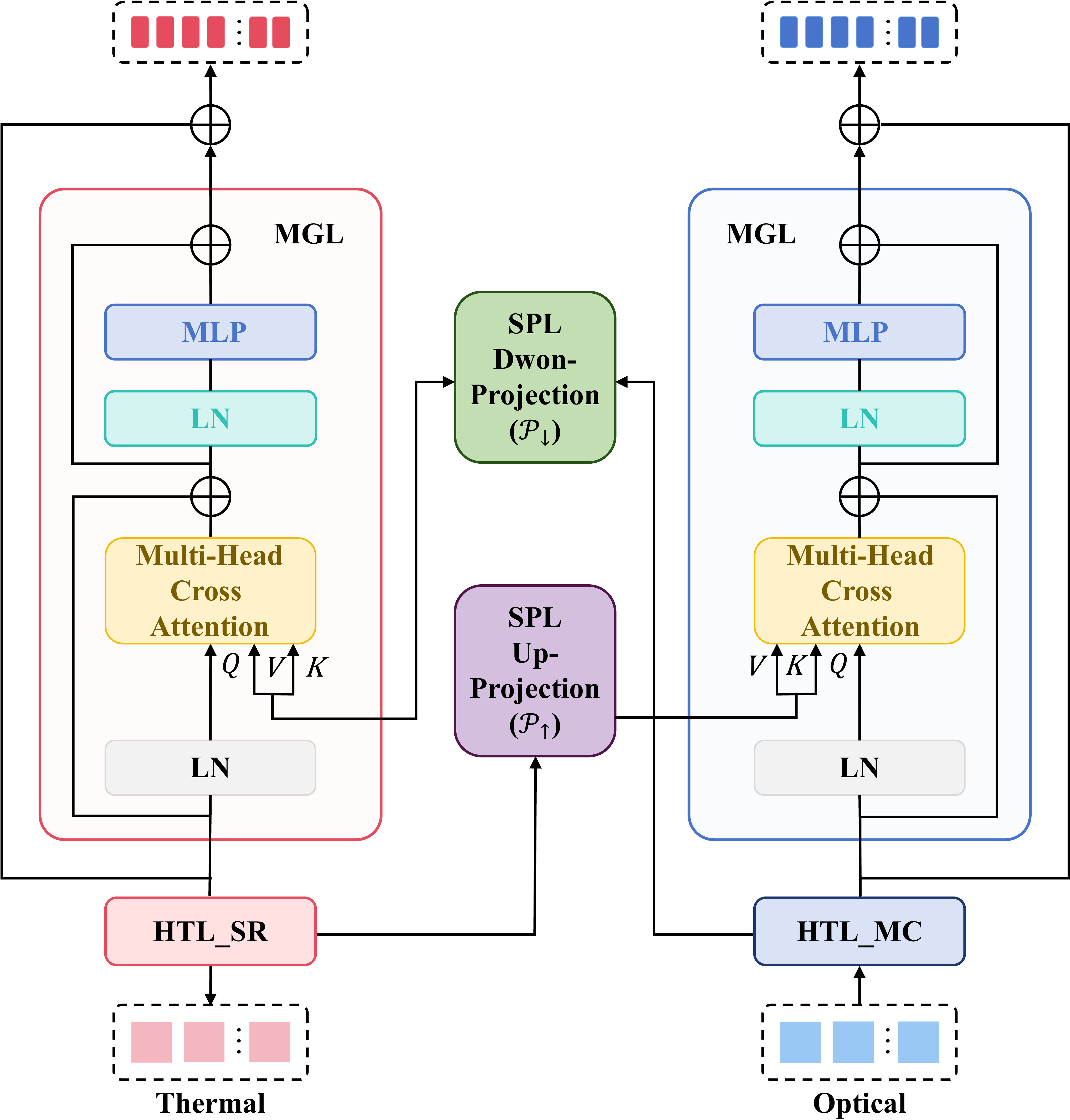}
    \caption{Framework of the proposed CRME, illustrating the SR branch (Left) and the MC branch (Right). Optical and thermal features are aligned via SPL and interact through MGL to mitigate domain gaps and resolution discrepancies.}
    \label{f_crme}
\end{figure}

\begin{figure}[t]
    \centering
    \includegraphics[width=0.85\linewidth]{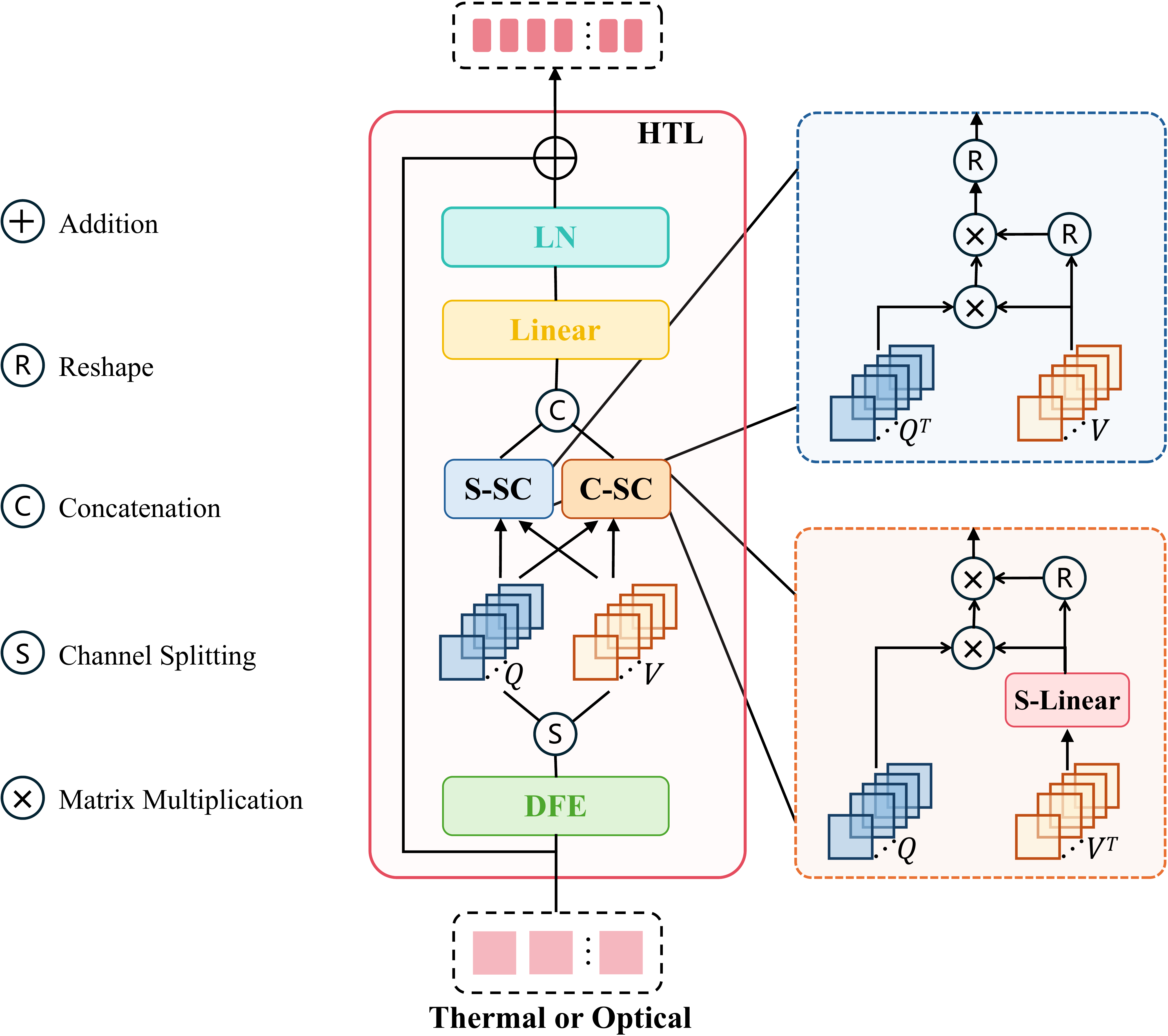}
    \caption{The network structure of HTL. Among these, spatial self-correlation (S-SC) and channel self-correlation (C-SC) aim to efficiently aggregate hierarchical information with computational complexity that scales linearly with window size.}
    \label{f_htl}
\end{figure}

\vspace{0.5em}
\noindent \textbf{Hierarchical Transformer Layer (HTL).}
To reduce computational overhead caused by the high resolution of optical patches, we adopted the lightweight HTL proposed in ~\cite{Zhang2024HITTransformer}, which experimentally demonstrated significant reductions in computational load and training time. The specific structure of HTL is shown in Fig.~\ref{f_htl}, which implements self-attention through two methods: spatial autocorrelation (S-SC) and channel autocorrelation (C-SC), both featuring linear computational complexity with respect to window size. 

\vspace{0.5em}
\noindent \textbf{Scale-Adaptive Projection Layer (SPL).}
Directly computing attention maps between multi-scale features is computationally inefficient and spatially misaligned. To address this, we design the SPL to dynamically project features into the target resolution, serving as the alignment basis for the subsequent attention mechanism. For the SR branch, we employ a down-projection operator $\mathcal{P}_{\downarrow}(\cdot)$ to condense high-frequency optical priors into the thermal LR space. Conversely, for the MC branch, an up-projection operator $\mathcal{P}_{\uparrow}(\cdot)$ maps thermal semantic structures to the optical HR space.

\vspace{0.5em}
\noindent \textbf{Mutual Guidance Attention Layer (MGL).}
Based on the aligned features from SPL, the MGL facilitates reciprocal information exchange. In this layer, the feature from the target domain serves as the Query ($Q$), while the projected feature from the auxiliary domain serves as the Key ($K$) and Value ($V$). 
For Optical-to-Thermal Guidance (SR Branch), the thermal feature queries the down-projected optical context:
\begin{equation}    
Q_T = \hat{F}_{T,i} \mathbf{W}_Q^T, \quad K_O = \mathcal{P}_{\downarrow}(\hat{F}_{O,i}) \mathbf{W}_K^T, \quad V_O = \mathcal{P}_{\downarrow}(\hat{F}_{O,i}) \mathbf{W}_V^T,
\end{equation}
where $\mathbf{W}^T$ are learnable projection matrices and $\hat{F}_{T,i}=f_{\mathrm{HTL}}(F_{T_{i-1}})$. The interactive thermal feature is then computed as:
\begin{equation}    
\tilde{F}_{T,i} = \hat{F}_{T,i} + \text{Softmax}\left(\frac{Q_T K_O^\top}{\sqrt{d_k}}\right) V_O.
\end{equation}
For Thermal-to-Optical Guidance (MC Branch), the optical feature queries the up-projected thermal semantics:
\begin{equation}    
Q_O = \hat{F}_{O,i} \mathbf{W}_Q^O, \quad K_T = \mathcal{P}_{\uparrow}(\hat{F}_{T,i}) \mathbf{W}_K^O, \quad V_T = \mathcal{P}_{\uparrow}(\hat{F}_{T,i}) \mathbf{W}_V^O.
\end{equation}
The spatially aligned thermal information is then propagated to the optical branch:
\begin{equation}     
\tilde{F}_{O,i} = \hat{F}_{O,i} + \text{Softmax}\left(\frac{Q_O K_T^\top}{\sqrt{d_k}}\right) V_T.
\end{equation}
By embedding the resolution alignment (SPL) and interaction (MGL) into a unified block, CRME effectively mitigates the domain gap and resolution discrepancy simultaneously.

\subsection{Physics-Driven Thermal Conduction Module}
\label{pdtm}
While CRME effectively transfers optical texture, unconstrained high-frequency details may introduce physical artifacts. To address this issue, the Physics-Driven Thermal Module (PDTM) is applied to constrain the thermal features $\tilde{F}_{T,i}$ following optical guidance. Thermal imaging is governed by heat conduction, where energy flows from high to low temperature regions until thermal equilibrium is reached, described by the heat equation:
\begin{equation}
\frac{\partial u}{\partial t} = \alpha \nabla^2 u = \alpha \left(\frac{\partial^2 u}{\partial x^2} + \frac{\partial^2 u}{\partial y^2}\right),
\end{equation}
where $\alpha$ denotes thermal diffusivity, which reflects the rate at which a material responds to heat conduction. 
At the macroscopic level, this microscopic thermal motion manifests as boundary blurring between objects. By integrating thermal conduction modeling~\cite{Chen2023Atmospheric,Chen2024Thermal3DGS}, we can understand the energy transfer patterns between objects, thereby enhancing sensitivity to temperature variations and improving the accuracy of thermal image super-resolution. As shown in Fig.~\ref{overall}(c), PDTM first extracts the second-order spatial derivatives of thermal images using the Laplacian operator to characterize the potential thermal diffusion field distribution:
\begin{equation}
\nabla^2 \tilde{F}_{T,i} = K_{Lap} \ast \tilde{F}_{T,i}.
\end{equation}
However, relying solely on the blurred edge features of degraded thermal images cannot simulate the actual heat conduction process. Therefore, we introduce optical images as geometric structure priors, first calculating the second-order gradients of thermal and optical features respectively, then normalizing them:
\begin{equation}
M_{T} = \frac{|\nabla^2 \tilde{F}_{T,i}|}{\max(|\nabla^2 \tilde{F}_{T,i}|) + \epsilon}, \quad
M_{O} = \frac{|\nabla^2 \tilde{F}_{O,i}|}{\max(|\nabla^2 \tilde{F}_{O,i}|) + \epsilon},
\end{equation}
where $\epsilon$ is a small constant to prevent division by zero. The guided Laplacian feature maps are constructed through weighted fusion:
\begin{equation}
M_{guided} = \lambda_{T} \cdot M_{T} + \lambda_{O} \cdot M_{O},
\end{equation}
where we empirically set $\lambda_{T} = \lambda_{O} = 0.5$. Subsequently, the fused edge features $M_{guided}$ are fed into a diffusivity predictor, which simulates the spatially-varying response of each pixel to thermal diffusion:
\begin{equation}
\mathbf{A} = f_{pre}(M_{guided}),
\end{equation}
To restore the thermal details governed by the predicted physical properties, we concatenate the learned diffusivity response $\mathbf{A}$ with the original thermal features $\tilde{F}_{T,i}$ along the channel dimension. The combined features are then fused through a convolution layer to generate the output of PDTM:
\begin{equation}
F_{Te,i} = \operatorname{Conv}_{out}([\tilde{F}_{T,i}, \mathbf{A}]) + \tilde{F}_{T,i}.
\end{equation}
This design effectively incorporates physical constraints into the deep network, enabling the model to adaptively learn accurate heat transfer processes and recover fine-grained thermal details. Finally, to allow each branch to focus on its respective task and extract deeper semantic features after interaction, the features ${F}_{Te,i}$ and $\tilde{F}_{O,i}$ are fed into $K$ cascaded Hierarchical Transformer Layers (HTL), producing the final output of the $i$-th stage:
\begin{equation}    
F_{T_{i}} = f_{HTL}^K({F}_{Te,i}) + F_{T_{i-1}}, \quad F_{O_{i}} = f_{HTL}^K(\tilde{F}_{O,i}) + F_{O_{i-1}}.
\end{equation}

\setlength{\floatsep}{5pt plus 2pt minus 2pt}
\setlength{\textfloatsep}{5pt plus 2pt minus 2pt}
\setlength{\intextsep}{5pt plus 2pt minus 2pt}
\subsection{Temperature Consistency Loss}
\label{loss}
In real-world thermal scenarios, due to thermal diffusion phenomena, the temperature distribution on an object's surface exhibits smooth and continuous variations, with abrupt changes occurring only at physical boundaries between materials with differing thermal properties. Within homogeneous regions, this smoothness manifests in thermal images as gradual temperature gradients with minimal abrupt discontinuities. Consequently, localized high-frequency variations in super-resolution outputs are often artifacts of model overfitting rather than genuine thermal information. To address this, we introduce a Temperature Consistency Loss achieved through two complementary physical priors as shown in Fig.~\ref{overall}(d): we use a pretrained Segment Anything Model (SAM) to extract region masks and boundary masks from the HR thermal images, and apply these masks to the HR and SR images to enforce regional distribution smoothness and boundary gradient consistency.

\vspace{0.5em}
\noindent\textbf{Regional distribution smoothness.} 
Within each semantic region, temperature values follow a consistent statistical distribution determined by material properties. Instead of pixel-level matching, we measure distributional divergence between super-resolved and ground-truth outputs. For each segmented instance $k$, we compute the 1-Wasserstein distance between normalized temperature histograms:
\begin{equation}
    \mathcal{L}_{region} = \frac{1}{K} \sum_{k=1}^{K} \| \operatorname{CDF}(h_{SR}^k) - \operatorname{CDF}(h_{HR}^k) \|_1,
\end{equation}
where $K$ denotes the total number of segmented instances, $h_{SR}^k$ and $h_{HR}^k$ represent the normalized temperature histograms of the $k$-th instance in the super-resolved and high-resolution images respectively, and $\operatorname{CDF}(\cdot)$ computes the cumulative distribution function.

\vspace{0.5em}
\noindent\textbf{Boundary gradient consistency.} 
To maintain strong thermal transitions at object boundaries, we enforce a gradient-based constraint. Using a binary boundary mask $M_b$ derived from instance segmentation, we penalize gradient deviations at region interfaces:
\begin{equation}
    \mathcal{L}_{bound} = \| (\nabla I_{SR} - \nabla I_{HR}) \odot M_{b} \|_1,
\end{equation}
where $I_{SR}$ and $I_{HR}$ denote the super-resolved and high-resolution thermal images respectively, $\nabla$ represents the gradient operator, $\odot$ denotes element-wise multiplication, and $M_b$ is a binary mask highlighting object boundaries. This prevents thermal bleeding artifacts and preserves physical discontinuities in multi-object scenes.
The overall training objective combines consistency and reconstruction losses as a weighted sum:
\begin{equation}
    \mathcal{L}_{total} = (1 - \lambda) \mathcal{L}_{rec} + \lambda \mathcal{L}_{tcl},
\end{equation}
where  $\mathcal{L}_{{rec}} = \|I_{\text{SR}} - I_{\text{HR}}\|_1$ represents the L1 reconstruction loss, $\mathcal{L}_{tcl} = \mathcal{L}_{region} + \mathcal{L}_{bound}$ combines the two consistency terms, and $\lambda \in [0,1]$ is a hyperparameter that balances reconstruction fidelity and thermophysical consistency.

\begin{table*}
\renewcommand{\arraystretch}{1.1}
\caption{Comparison with state-of-the-art SISR methods and GISR methods for $\times$4 and $\times$8 SR on the VGTSR2.0 dataset, employing BI and BD degradation models. \textbf{Best} and \underline{\textit{Second best}}. Higher PSNR and SSIM are better. \textsuperscript{*} indicates methods adapted by us for the GISR task.}
\label{tab:vgtsr2.0-experiment}
\begin{tabular}{lcccccccccc}
\toprule
\multirow{2}{*}{Method} & \multirow{2}{*}{Venue} & \multirow{2}{*}{G/S} & \multicolumn{2}{c}{4BI} & \multicolumn{2}{c}{4BD} & \multicolumn{2}{c}{8BI} & \multicolumn{2}{c}{8BD}\\
\cmidrule{4-7} \cmidrule{8-11}
& & & PSNR$\uparrow$ & SSIM$\uparrow$ & PSNR$\uparrow$ & SSIM$\uparrow$ & PSNR$\uparrow$ & SSIM$\uparrow$ & PSNR$\uparrow$ & SSIM$\uparrow$ \\
\hline
Bicubic &-- &Single &26.76 &0.8054 &25.77 &0.7833 &22.74 &0.6364 &22.44 &0.6485 \\
EDSR\cite{Lim2017EDSR}&CVPR'2017 &Single &30.57 &0.8997 &30.38 &0.8978 &24.55 &0.7406 &24.54 &0.7462 \\
RCAN\cite{Zhang2018RCAN}&CVPR'2018 &Single &31.26 &0.9106 &31.06 &0.9080 &25.62 &0.7657 &24.75 &0.7544\\
SAN\cite{Dai2019SAN}&CVPR'2019 &Single &31.20 &0.9099 &31.09 &0.9084 &25.63 &0.7581 &24.76 &0.7649 \\
HAN\cite{Niu2020HAN}&ECCV'2020 &Single &31.23 &0.9080 &30.81 &0.9041 &25.60 &0.7588 &24.78 &0.7595 \\
SwinIR\cite{Liang2021SwinIR}&ICCV'2021 &Single &31.19 &0.9088 &31.06 &0.9089 &25.70 &0.7663 &24.86 &0.7618\\
Restormer\cite{Zamir2022Restormer}&CVPR'2022 &Single &30.99 &0.9053 &30.84 &0.9050 &25.73 &0.7677 &24.90 &0.7615\\
HAT\cite{Chen2023HAT}&CVPR'2023 &Single &31.26 &0.9086 &31.10 &0.9079 &25.72 &0.7662 &24.89 &0.7616\\
SPT\cite{Hao2024SPT}&TGSR'2024 &Single &31.01 &0.9058 &30.83 &0.9046 &25.66 &0.7652 &24.84 &0.7589\\
HIT\cite{Zhang2024HITTransformer}&ECCV'2024 &Single &30.85 &0.9043 &30.74 &0.9040 &25.54 &0.7618 &24.76 &0.7581\\
UGSR\cite{Gupta2021UGSR}&TIP'2021 &Guided &30.53 &0.8972 &30.34 &0.8962 &25.35 &0.7536 &24.64 &0.7483\\
MGNet\cite{Zhao2023MGNet}&TGSR'2023 &Guided &31.33 &0.9116 &31.16 &0.9100 &26.12 &0.7829 &25.41 &0.7803 \\
SwinFuSR\cite{Arnold2024SwinFuSR}&CVPR'2024 &Guided &28.95 &0.8652 &28.89 &0.8662 &24.45 &0.7214 &23.91 &0.7157 \\
CENet\cite{Zhao2024CENet}&TGSR'2024 &Guided &31.42 &0.9146 &31.26 &0.9120 &26.28 &0.7877 &25.50 &0.7871 \\ 
GDNet\cite{ZHAO2025GDNet}&ISPRS'2025 &Guided &\underline{31.51} &\underline{0.9159} &\underline{31.40} &\underline{0.9134} &\underline{26.38} &\underline{0.7930} &\underline{25.67} &\underline{0.7937} \\
CATANet\textsuperscript{*} \cite{Liu2025CATANet}&CVPR'2025 &Guided &30.77	&0.9000 &30.50 	&0.8995	&25.62 	&0.7659 &24.18	&0.7371  \\
DuCos\cite{Yan2025DUCOS} &ICCV'2025 &Guided &31.27 &0.9098 &30.30  &0.8959  &24.61 &0.7393 &23.94 &0.7292  \\
\hline
PCNet(Ours) &-- & Guided &\textbf{31.76} &\textbf{0.9200} &\textbf{31.55} &\textbf{0.9174} &\textbf{26.54}&\textbf{0.8003} &\textbf{25.79} &\textbf{0.7946} \\
\bottomrule
\end{tabular}
\end{table*}

\section{Experiments}
This section presents a comprehensive evaluation of the proposed PCNet along with comparisons to state-of-the-art methods on the VGTSR2.0 and DroneVehicle datasets. We begin by introducing the optical-thermal image datasets captured from UAV perspectives and the primary evaluation metrics used in Section~\ref{Datasets}. Subsequently, we detail the implementation of PCNet in Section~\ref{Implementation}. The effectiveness of PCNet in GISR and downstream tasks is analyzed in Section~\ref{Quantitative1} and Section~\ref{Quantitative2}, respectively. Finally, ablation studies and visualizations are reported in Section~\ref{Ablations}.

\subsection{Datasets and Evaluation Metrics}
\label{Datasets}
The VGTSR2.0 dataset is a large-scale multi-modal benchmark specifically designed for optics-guided thermal UAV image SR. It comprises 3,500 manually registered and aligned optical-thermal image pairs uniformly normalized to $640 \times 512$ resolution, partitioned into 2,798 training pairs and 702 test pairs. The dataset encompasses eight typical scenarios such as single-story buildings, skyscrapers, parking lots, farmland, ponds, lakes, streets, and schools, captured under diverse environmental conditions (Normal, Low-light, and Fog), effectively simulating real-world UAV application scenarios.

The DroneVehicle dataset is a large-scale optical-thermal vehicle detection dataset captured by a UAV platform, comprising 28,439 pre-aligned image pairs at $840 \times 712$ resolution across diverse scenarios including urban areas, roads, residential zones, and parking lots. Due to its scale, we strategically sampled 1,000 training and 300 test images with temporal diversity to ensure independent data distributions. To adapt to the optics-guided thermal image SR task, we first employed an algorithm to remove annotated boundaries from the images and uniformly resize them to a resolution of $640 \times 512$. Subsequently, low-resolution inputs were generated from the HR thermal images through bicubic (BI) and blur-downscale (BD) degradation models to simulate realistic degradation patterns.

To quantitatively evaluate the quality of generated HR thermal images, we employ two widely adopted metrics: Peak Signal-to-Noise Ratio (PSNR) and Structural Similarity Index (SSIM). Notably, PSNR is computed directly on all channels provided by the dataset rather than the traditional luminance (Y) component in the YCbCr color space, enabling a more comprehensive assessment of color fidelity across all spectral channels. Higher values of PSNR and SSIM indicate superior image quality.

\subsection{Implementation Details}
\label{Implementation}
We implement PCNet using PyTorch on the RTX 3090 GPU platform. We use $640\times512$ optical images as inputs for the MC branch, while the SR branch takes low-resolution thermal images generated by the BI and BD degradation models. In our proposed PCNet, the size of the convolution kernel, embedding dimension, window size, number of attention heads, and patch size are generally set to $3 \times 3$, 96, 8, 6, 1, respectively. Empirically, the best performance is achieved with the temperature consistency loss weight $\lambda$ set to 0.03. Our model is configured with $N=4$ cascaded processing stages. Each stage comprises a Cross-Resolution Mutual Enhancement Module (CRME) in each branch, a Physics-Driven Thermal Conduction Module (PDTM) shared across branches, and six Hierarchical Transformer Layers (HTL) per branch ($K=6$). Following common practice, we randomly crop images during the training phase, using the resulting patches as model inputs, while full images are fed into the model during testing. To be consistent with the baseline SwinIR and other SR methods, we crop each image to obtain a patch with a pixel size of $48 \times 48$ as the input. For both $\times$ 4 and $\times$8 SR tasks, LR thermal images are cropped into $48 \times 48$ pixel patches. The HR optical images are cropped into $192 \times 192$ pixel patches for $\times$4 SR and $384 \times 384$ pixel patches for $\times$8 SR from the corresponding positions. We employ the Adam optimizer ($\beta_{1}$ = 0.9, $\beta_{2}$ = 0.99, and $\epsilon$ = $10^{-8}$) and batch size of 8 for end-to-end training. The learning rate is initially set to $10^{-4}$ and is then reduced by half every 200 epochs.

\begin{table*}
\renewcommand{\arraystretch}{1.1}
\caption{Comparison with state-of-the-art SISR methods and GISR methods for $\times$4 and $\times$8 SR on the DroneVehicle dataset, employing BI and BD degradation models. \textbf{Best} and \underline{\textit{Second best}}. Higher PSNR and SSIM are better. \textsuperscript{*} indicates methods adapted by us for the GISR task.}
\label{tab:droneVehicle-experiment}
\begin{tabular}{lcccccccccc}
\toprule
\multirow{2}{*}{Method} & \multirow{2}{*}{Venue} & \multirow{2}{*}{G/S} & \multicolumn{2}{c}{4BI} & \multicolumn{2}{c}{4BD} & \multicolumn{2}{c}{8BI} & \multicolumn{2}{c}{8BD}\\
\cmidrule{4-7} \cmidrule{8-11}
& & & PSNR$\uparrow$ & SSIM$\uparrow$ & PSNR$\uparrow$ & SSIM$\uparrow$ & PSNR$\uparrow$ & SSIM$\uparrow$ & PSNR$\uparrow$ & SSIM$\uparrow$ \\
\hline
Bicubic &-- &Single &25.25 &0.7806 &24.36 &0.7068 &21.10 &0.6088 &20.51 &0.6129 \\
EDSR\cite{Lim2017EDSR}&CVPR'2017 &Single &29.49 &0.8685 &26.38 &0.8345 &22.44 &0.6781 &22.14 &0.6913 \\
RCAN\cite{Zhang2018RCAN}&CVPR'2018 &Single &29.71 &0.8709 &26.79 &0.8361 &22.48 &0.6835 &22.35 &0.6993\\
SAN\cite{Dai2019SAN}&CVPR'2019 &Single &29.71 &0.8710 &27.02 &0.8397 &22.52 &0.6844 &22.25 &0.6986 \\
HAN\cite{Niu2020HAN}&ECCV'2020 &Single &29.45 &0.8671 &26.91 &0.8402 &22.52 &0.6820 &22.27 &0.6979 \\
SwinIR\cite{Liang2021SwinIR}&ICCV'2021 &Single &29.78 &0.8720 &26.41 &0.8336 &22.53 &0.6856 &22.09 &0.6956\\
Restormer\cite{Zamir2022Restormer}&CVPR'2022 &Single &29.73 &0.8712 &26.93 &0.8356 &22.52 &0.6852 &22.06 &0.6949\\
HAT\cite{Chen2023HAT}&CVPR'2023 &Single &29.80 &0.8725 &26.63 &0.8325 &22.68 &0.6865 &22.23 &0.6979\\
SPT\cite{Hao2024SPT}&TGSR'2024 &Single &29.60 &0.8697 &27.10 &0.8404 &22.37 &0.6766 &21.95 &0.6887\\
HIT\cite{Zhang2024HITTransformer}&ECCV'2024 &Single &29.66 &0.8705 &26.50 &0.8348 &22.52 &0.6841 &22.43 &0.7010\\
UGSR\cite{Gupta2021UGSR}&TIP'2021 &Guided &30.44 &0.8802 &28.50 &0.8623 &24.83 &0.7619 &23.79 &0.7514\\
MGNet\cite{Zhao2023MGNet}&TGSR'2023 &Guided &31.45 &0.8907 &29.22 &0.8744 &26.53 &0.8023 &25.92 &0.8021 \\
SwinFuSR\cite{Arnold2024SwinFuSR}&CVPRW'2024 &Guided &30.85 &0.8852 &28.96 &0.8696 &24.88 &0.7527 &24.43 &0.7532 \\
CENet\cite{Zhao2024CENet}&TGSR'2024 &Guided &\underline{31.61} &\underline{0.8923} &\underline{29.99} &\underline{0.8799} &\underline{26.86} &\underline{0.8090} &\underline{26.69} &\underline{0.8146} \\ 
CATANet\textsuperscript{*} \cite{Liu2025CATANet}&CVPR'2025 &Guided &29.65 &0.8697 &25.14  &0.8078  &22.57  &0.6858  &22.73  &0.7091  \\
DuCos\cite{Yan2025DUCOS} &ICCV'2025 &Guided &30.77 &0.8839 &28.22 &0.8600 &25.21  &0.7784  &24.94 &0.7811 \\
\hline
PCNet(Ours) &-- & Guided &\textbf{31.76} &\textbf{0.8936} &\textbf{30.22} &\textbf{0.8820} &\textbf{27.05}&\textbf{0.8123} &\textbf{26.80} &\textbf{0.8163} \\
\bottomrule
\end{tabular}
\end{table*}

\begin{table}
\centering
\renewcommand{\arraystretch}{1.5}
\caption{Quantitative comparison on the VGTSR2.0 dataset no-reference evaluation metrics NIQE and BRISQUE, for $\times$4 SR employing the BI degradation model. \textbf{Best} and \underline{\textit{Second best}}. Lower NIQE and BRISQUE are better. \textsuperscript{*} indicates methods adapted by us for the GISR task.}
\label{tab:4BI-no-reference}
\scriptsize 
\begin{tabular}{lcc|ccc}
\toprule
Method        & NIQE $\downarrow$ & BRISQUE $\downarrow$ &Param & FLOPs &Time  \\ \hline
Bicubic      & 8.0723            & 60.1769                       &-      & - & -        \\
EDSR\cite{Lim2017EDSR}         & 6.0295            & 48.6591                      &43M      &10.22G  &0.0020s       \\
RCAN\cite{Zhang2018RCAN}         & 6.1085            & 47.4327                       &16M      &67.21G   &0.0422s       \\
SAN\cite{Dai2019SAN}          & 6.0908            & 47.0781                      &15.7M      &67.00G  &0.0286s        \\
HAN\cite{Niu2020HAN}          & 6.0651            & 47.3728                       &15.4M      &123.85G   & 0.0481s       \\
SwinIR\cite{Liang2021SwinIR}       & 6.0226            & 47.1836                       &5M      &12.56G   & 0.0213s        \\
Restormer\cite{Zamir2022Restormer}    & 6.0493            & 47.3076                       &25.3M      &23.79G  &0.0398s      \\
HAT\cite{Chen2023HAT}          & 6.0241            & 47.3337                       &5.2M      &13.18G    & 0.0414s     \\
SPT\cite{Hao2024SPT}          & 6.0928            & 47.4051                  &3.2M      &8.18G         & 0.0260s \\
HIT\cite{Zhang2024HITTransformer}     & 6.0445   & 48.1053     & 2.0M   & 8.62G   & 0.0198s    \\
UGSR\cite{Gupta2021UGSR}         & 5.9879            & 48.0676                       &4.5M      &46.60G     & 0.0029s     \\
MGNet\cite{Zhao2023MGNet}        & 5.9544            & 46.7480                       &18.6M      &49.41G     & 0.0664s    \\
SwinFuSR\cite{Arnold2024SwinFuSR}     &6.5776    &47.8350      & 4.8M   & 20.33G   & 0.0279s    \\
CENet\cite{Zhao2024CENet}        & 5.9764            & 45.8960                       &11.8M      &28.89G  & 0.0486s        \\
GDNet\cite{ZHAO2025GDNet}  & \underline{5.9226}    & 46.6237               &11.9M     &32.07G    &0.0531s      \\ 
CATANet\textsuperscript{*} \cite{Liu2025CATANet}     & 5.9333   & \underline{45.5147}     & 3.5M   & 16.61G   & 0.0211s    \\
DuCos\cite{Yan2025DUCOS}    &6.2490      &47.3806    & 3.4M   &39.2G   &0.0369s     \\ \hline
PCNet(ours)     & \textbf{5.9028}   & \textbf{45.3868}     & \textbf{8.8M}   & \textbf{48.86G}   & \textbf{0.0613s}    \\  \bottomrule
\end{tabular}
\end{table}

\subsection{Quantitative and Qualitative Evaluation}
\label{Quantitative1}
To validate the effectiveness of PCNet, we conduct $\times$4 and $\times$8 SISR and GISR experiments on the VGTSR2.0 dataset and the DroneVehicle dataset. We compare our method against nine state-of-the-art SISR methods, including CNN-based models EDSR~\cite{Lim2017EDSR}, RCAN~\cite{Zhang2018RCAN}, SAN~\cite{Dai2019SAN}, and HAN~\cite{Niu2020HAN}, as well as Transformer-based models SwinIR~\cite{Liang2021SwinIR}, Restormer~\cite{Zamir2022Restormer}, HAT~\cite{Chen2023HAT}, SPT~\cite{Hao2024SPT}, and HIT~\cite{Zhang2024HITTransformer}. Additionally, we evaluate against seven guided SR methods: UGSR~\cite{Gupta2021UGSR}, MGNet~\cite{Zhao2023MGNet}, SwinFuSR~\cite{Arnold2024SwinFuSR}, CENet~\cite{Zhao2024CENet}, CATANet~\cite{Liu2025CATANet}, DuCos~\cite{Yan2025DUCOS}, and GDNet~\cite{ZHAO2025GDNet}, where GDNet currently represents the state-of-the-art performance.

Table~\ref{tab:vgtsr2.0-experiment} presents the results for $\times$4 and $\times$8 SR employing BI and BD degradation models on the VGTSR2.0 dataset. Compared to SISR and GISR methods, for $\times$4 SR, our method achieves a 0.25 dB improvement in PSNR and a 0.0041 increase in SSIM when using the BI model, and a 0.15 dB and 0.0040 improvement with the BD model. For $\times$8 SR, the BI model yields a 0.16 dB PSNR and 0.0073 SSIM enhancement, while the BD model achieves a 0.12 dB PSNR and a 0.0009 SSIM increase. The different SISR methods exhibit similar performance levels, while Transformer-based SR methods generally outperform those based on CNNs. In the GISR methods, although UGSR, SwinFuSR, and DuCos utilize HR optical images as guidance, their relatively shallow network architectures and simplistic feature fusion approaches limit the full utilization of optical features, instead introducing redundant information and noise detrimental to thermal image reconstruction, resulting in inferior PSNR and SSIM values. MGNet outperforms nearly all SISR methods because it extracts clues such as edges, semantics, and appearance from optical images, thereby reducing the impact of noise. CENet and GDNet respectively incorporate MC and task-assisted SR, along with multi-scenario optical image-guided disentanglement and adaptive fusion strategies, thereby offering significant advantages over other methods. Our proposed PCNet achieves optimal performance by employing more high-frequency optical priors through cross-resolution mutual enhancement and utilizing thermal conduction to physically constrain the optical guidance process.

\begin{figure*}
    \centering
    \includegraphics[width=0.95\linewidth]{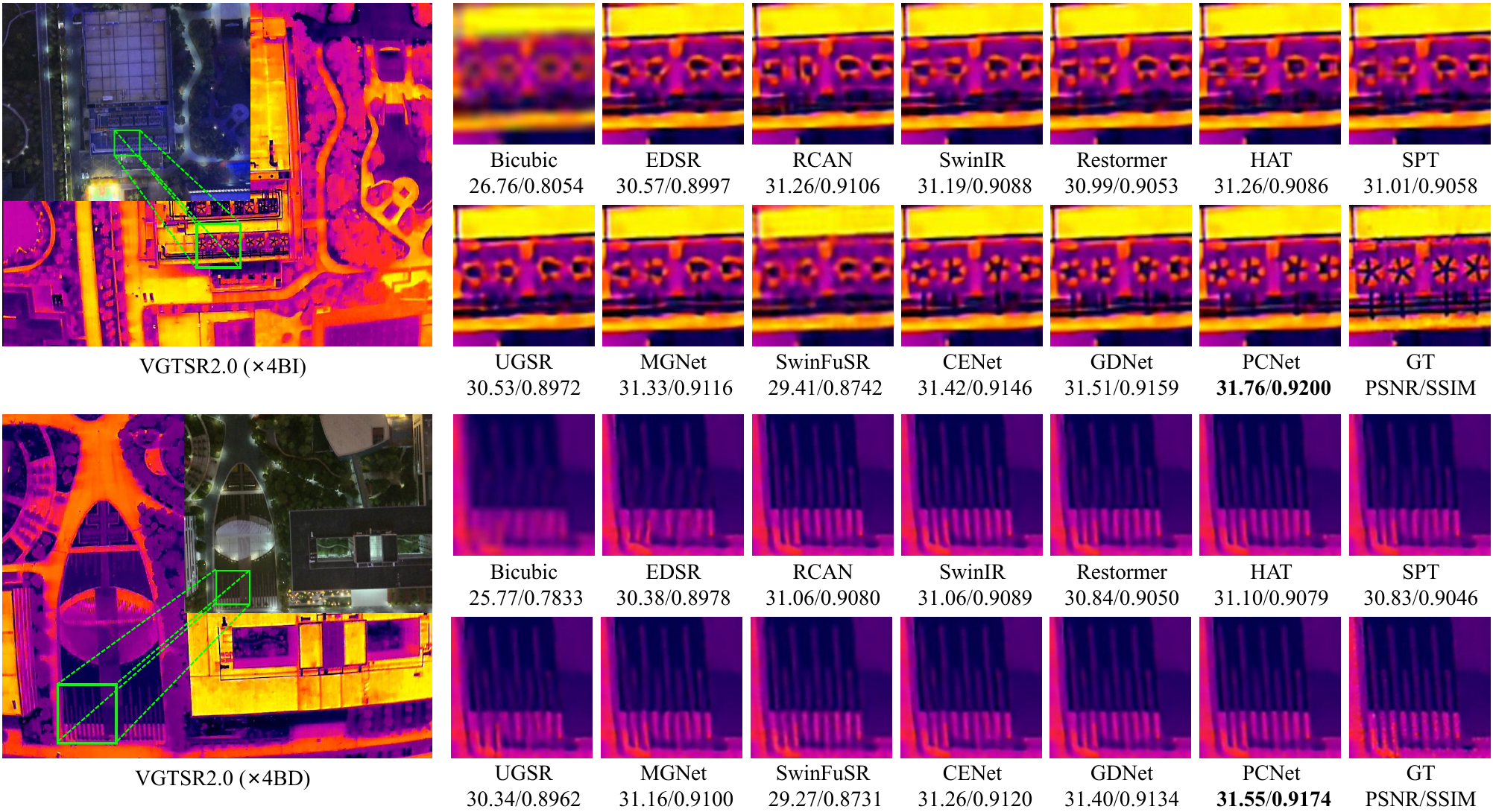}
    \caption{Qualitative comparison of super-resolution methods at $\times$4 SR with BI and BD degradation models in the VGTSR2.0 dataset. Experimental results demonstrate that our proposed PCNet effectively utilizes information from optical images to recover texture details and clear edges closer to the GT.}
    \label{vgtsr2.0_x4}
\end{figure*}
\begin{figure*}
    \centering
    \includegraphics[width=0.95\linewidth]{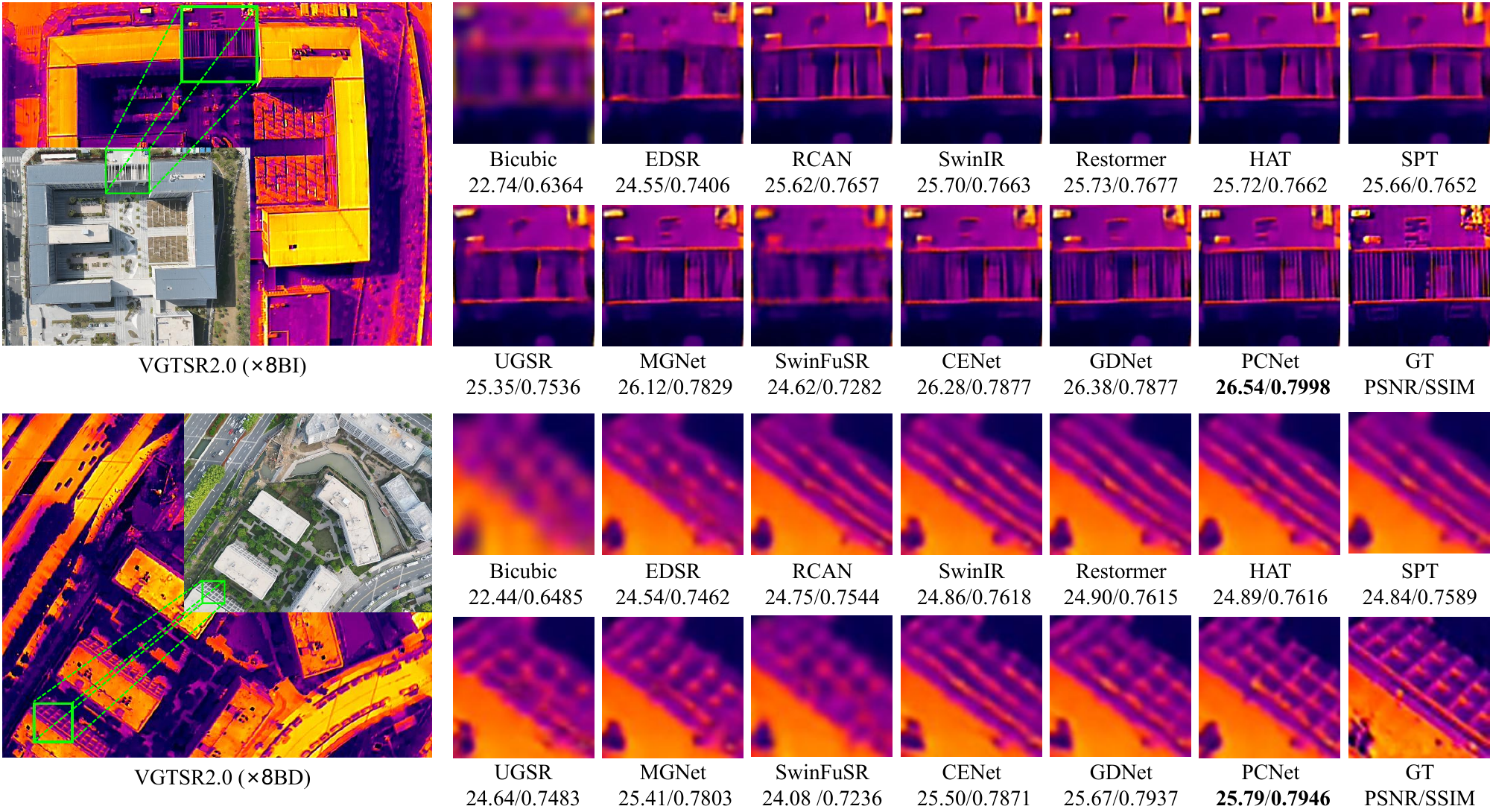}
    \caption{Qualitative comparison of super-resolution methods at $\times$8 SR with BI and BD degradation models in the VGTSR2.0 dataset. Our proposed PCNet can more accurately restore array/grid structures and local hotspot shapes while maintaining high local contrast and edge sharpness, outperforming most existing methods.}
  \label{vgtsr2.0_x8}
\end{figure*}

\begin{figure*}    
\centering    
\includegraphics[width=0.95\linewidth]{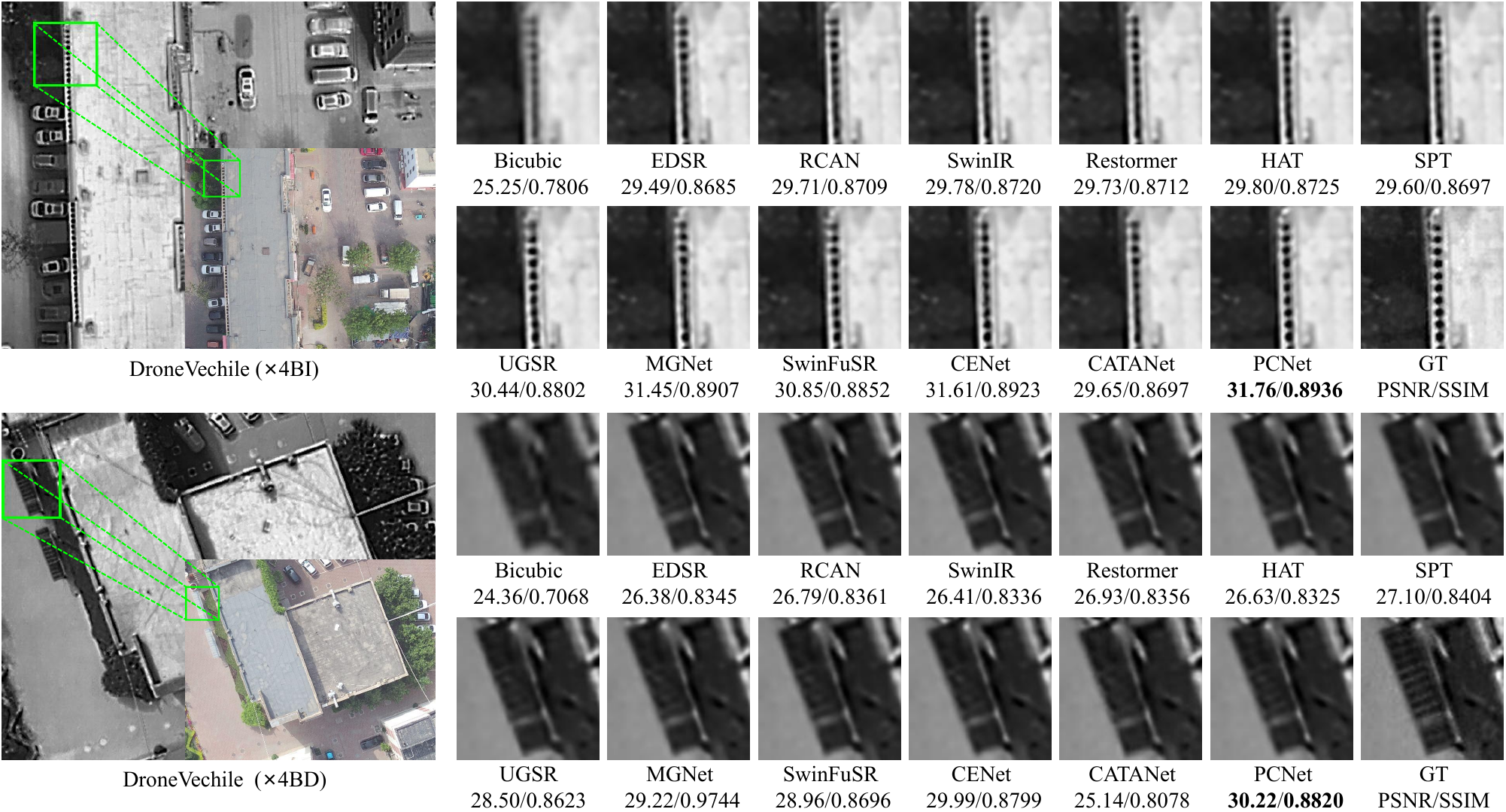}    
\caption{Qualitative comparison of super-resolution methods at $\times$4 SR with BI and BD degradation models in the DroneVehicle dataset. Compared to advanced SISR and GISR methods including MGNet and CENet, our method yields reconstruction results that are closer to the GT.}    
\label{droneVehicle_x4}
\end{figure*}

\begin{figure*}    
\centering    
\includegraphics[width=0.95\linewidth]{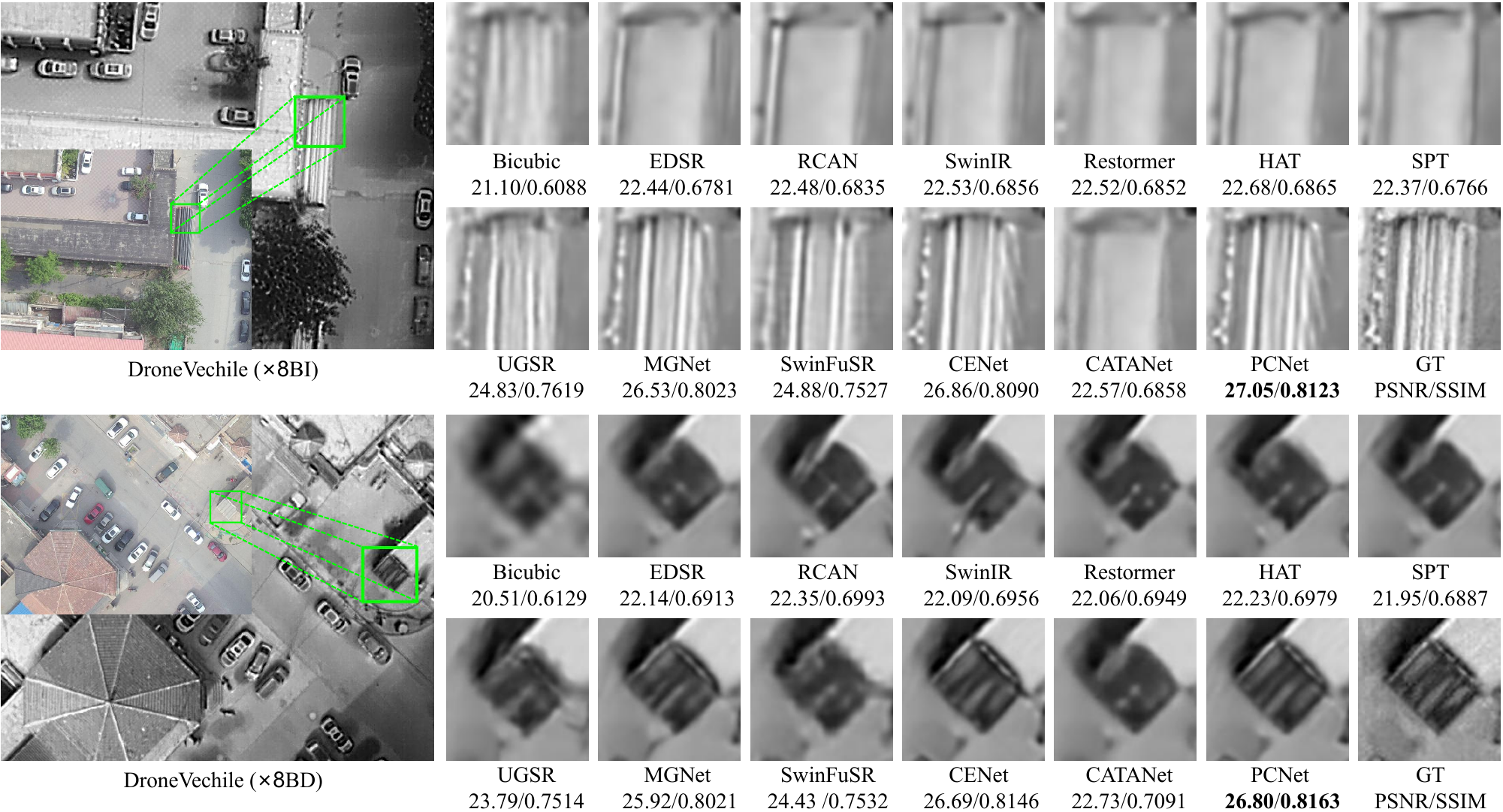}    
\caption{Qualitative comparison of super-resolution methods at $\times$8 SR with BI and BD degradation models in the DroneVehicle dataset. Our method also achieves better results on the $\times$8 degradation model.}  
\label{droneVehicle_x8}
\end{figure*}

The quantitative results in Table~\ref{tab:droneVehicle-experiment} also validate the superiority of our method on the DroneVehicle dataset. Compared to existing methods, for $\times$4 SR, our PCNet achieves a 0.15 dB improvement and a 0.23 dB increase in PSNR when employing the BI and the BD degradation model, respectively. For $\times$8 SR, the PSNR improves by 0.19 dB in the BI degradation model and by 0.11 dB with the BD degradation model, underscoring the effective performance of PCNet in optics-guided thermal SR. To further validate the effectiveness of our method in mitigating artifacts and image distortion, we conduct a no-reference image quality assessment using NIQE\cite{Mittal2012NIQE} and BRISQUE\cite{Mittal2012BRISQUE} for $\times$4 SR methods on the VGTSR2.0 dataset, with the quantitative results summarized in Table~\ref{tab:4BI-no-reference}. The experimental results demonstrate that our PCNet significantly improves performance in both NIQE and BRISQUE metrics compared to other methods. Table~\ref{tab:4BI-no-reference} also presents a comparison of different SR methods based on model size, FLOPs, and inference performance, with our model demonstrates advantages in terms of parameter efficiency.  Compared to MGNet and GDNet, our model reduces the parameter count by 53\% and 26\%, which indicates that our approach not only reduces the parameter size compared to the state-of-the-art model GDNet but also achieves outstanding performance. 

\begin{figure*}
    \centering
    \includegraphics[width=1.0\linewidth]{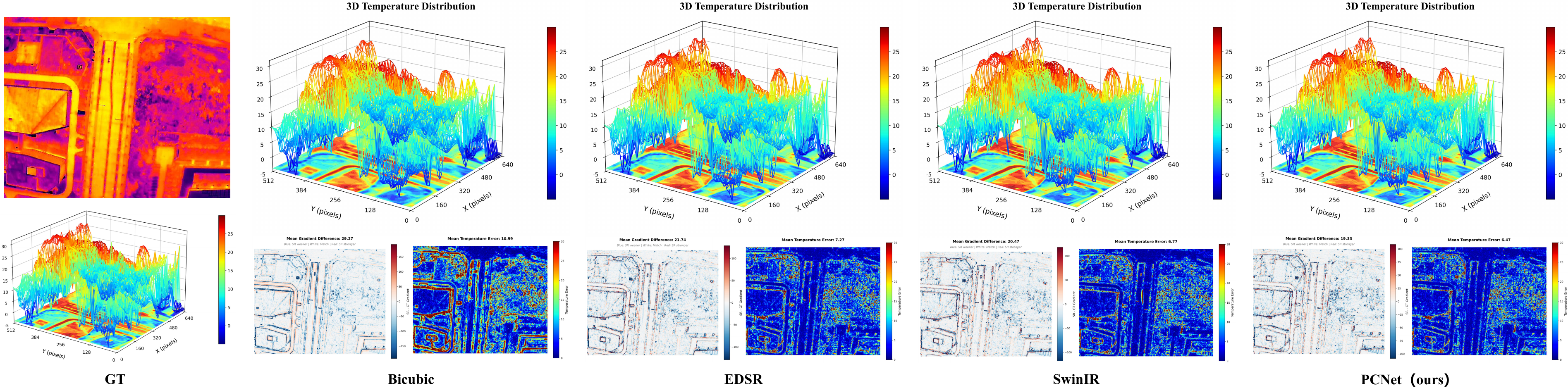}
    \caption{Visual comparison showing 3D temperature distribution, temperature difference, and gradient difference for different SR methods compared to GT.}
    \label{temperature}
\end{figure*}

\begin{figure*}
    \centering
    \includegraphics[width=0.95\linewidth]{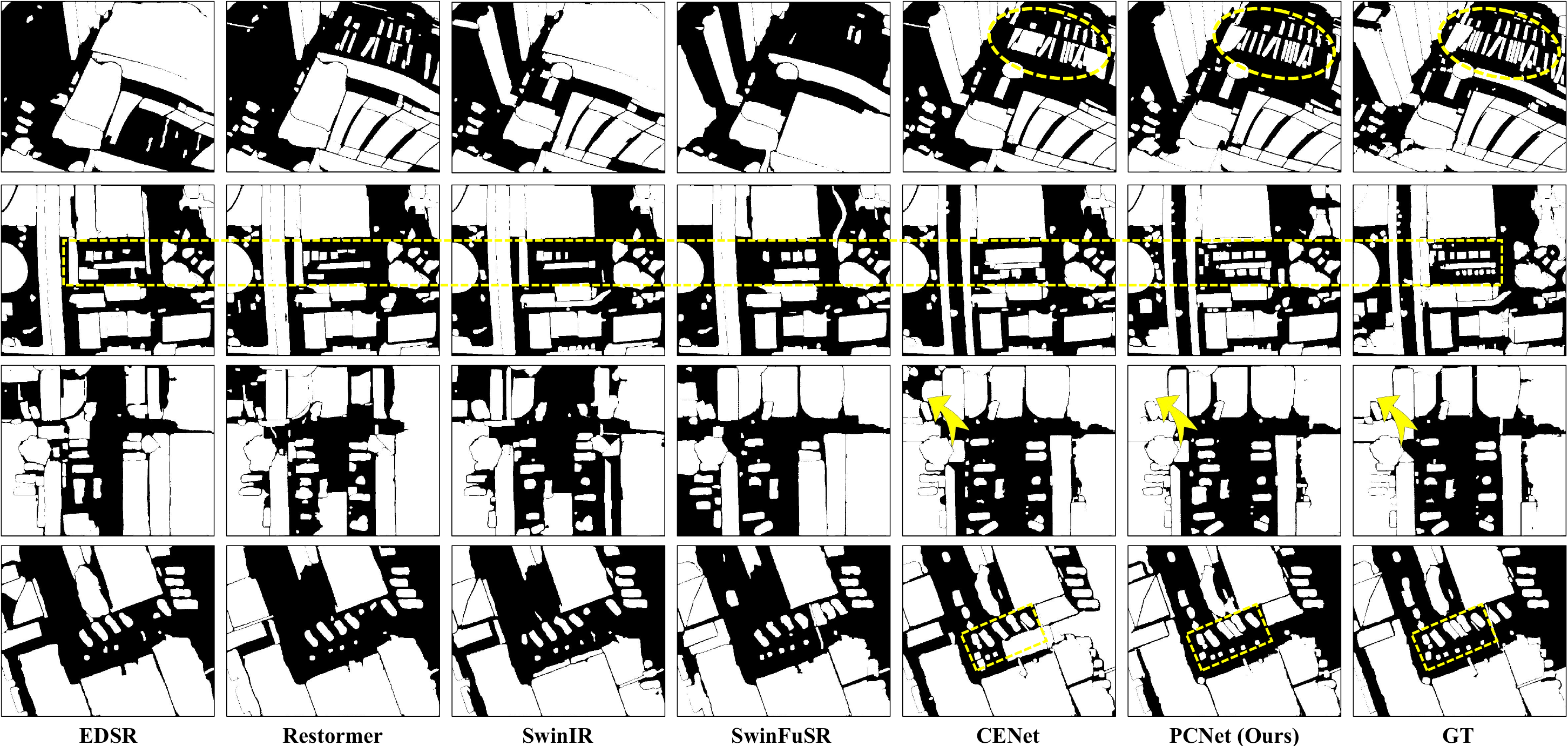}
    \caption{Visual comparison of segmentation results after SR using different methods. Comparison indicates that the reconstruction results from PCNet outperform those of other methods in downstream segmentation tasks.}
    \label{segmentation}
\end{figure*}

\begin{table}[t]
\centering
\renewcommand{\arraystretch}{1.3}
\scriptsize 
\caption{Quantitative comparison of segmentation results on VGTSR2.0 dataset. \textbf{Best} and \underline{\textit{Second best}} (SR Scale Factor 8).}
\label{tab:seg_vgtsr}
\begin{tabular}{c|cccc}
\hline
Method & IoU (\%) $\uparrow$ & Dice (\%) $\uparrow$ & ACC (\%) $\uparrow$ & ASSD $\downarrow$ \\
\hline
EDSR      & 65.92 & 77.01 & 78.28 & 15.50 \\
Restormer & 66.10 & 77.38 & 79.28 & 13.58 \\
SwinIR    & 66.20 & 77.50 & 79.17 & 14.40 \\
SwinFuSR  & 62.67 & 73.97 & 70.67 & 29.92 \\
CENet     & \underline{69.29} & \underline{79.90} & \underline{81.22} & \underline{12.93} \\
\hline
PCNet (Ours) & \textbf{69.54} & \textbf{80.02} & \textbf{81.50} & \textbf{12.21} \\
\hline
\end{tabular}
\end{table}

Our method not only demonstrates superior performance on quantitative evaluation metrics such as PSNR and SSIM, but also yields significant improvements in perceptual quality. As illustrated in Figs.~\ref{vgtsr2.0_x4}, ~\ref{vgtsr2.0_x8}, ~\ref{droneVehicle_x4}, and ~\ref{droneVehicle_x8}, SISR methods struggle to recover detailed textures and suffer from severe blurring artifacts, particularly in large-scale SR. While GISR approaches are capable of preserving structural information guided by optical images, the significant discrepancy between the two modalities inevitably introduces noise from optical images into the thermal counterparts, which leads to texture clutter and structural distortion. In contrast, our proposed PCNet leverages the sufficient guidance of optical information under physical constraints to reconstruct texture details and clear edges that are physically consistent with thermal images.

\begin{table}[t]
\centering
\renewcommand{\arraystretch}{1.3}
\scriptsize 
\caption{Quantitative comparison of segmentation results on DroneVehicle dataset. \textbf{Best} and \underline{\textit{Second best}} (SR Scale Factor 8).}
\label{tab:seg_drone}
\begin{tabular}{c|cccc}
\hline
Method & IoU (\%) $\uparrow$ & Dice (\%) $\uparrow$ & ACC (\%) $\uparrow$ & ASSD $\downarrow$ \\
\hline
EDSR      & 55.90 & 70.12 & 74.96 & 11.42 \\
Restormer & 58.14 & 71.85 & 76.67 & 10.32 \\
SwinIR    & 58.03 & 71.66 & 76.42 & 10.52 \\
SwinFuSR  & 62.35 & 75.35 & 79.83 & 9.19 \\
CENet     & \underline{64.94} & \underline{77.35} & \textbf{80.78} & \textbf{8.03} \\
\hline
PCNet (Ours) & \textbf{65.21} & \textbf{77.36} & \underline{80.73} & \underline{8.29} \\
\hline
\end{tabular}
\end{table}

\begin{table*}[t]
\centering
\renewcommand{\arraystretch}{1.3}
\scriptsize 
\caption{Quantitative comparison of object detection results on the VGTSR2.0 and DroneVehicle datasets. \textbf{Best} and \underline{\textit{Second best}} (SR Scale Factor 8).}
\label{tab:detection}
\begin{tabular}{c|cccc|cccc}
\hline
\multirow{2}{*}{Method} & \multicolumn{4}{c|}{VGTSR2.0} & \multicolumn{4}{c}{DroneVehicle} \\
\cline{2-9}
 & mAP$_{50}$ (\%) $\uparrow$ & mAP (\%) $\uparrow$ & Recall (\%) $\uparrow$ & F1 (\%) $\uparrow$ & mAP$_{50}$ (\%) $\uparrow$ & mAP (\%) $\uparrow$ & Recall (\%) $\uparrow$ & F1 (\%) $\uparrow$ \\
\hline
EDSR & 33.31 & 25.76 & 4.21 & 7.90 & 85.90 & 74.73 & 73.43 & 83.08 \\
Restormer & 43.01 & 32.63 & 10.84 & 18.95 & 87.80 & 77.07 & 76.97 & 85.43 \\
SwinIR    & 40.59 & 30.71 & 9.21 & 16.34 & 87.96 & 77.03 & 77.50 & 85.55 \\
SwinFuSR  & 27.99 & 21.35 & 4.33 & 8.00 & 90.63 & 82.89 & 81.93 & 89.04 \\
CENet     & \underline{44.66} & \underline{35.49} & \underline{11.80} & \underline{20.48} & \underline{93.93} & \underline{90.50} & \underline{88.55} & \underline{92.25} \\
\hline
PCNet (Ours) & \textbf{49.60} & \textbf{40.14} & \textbf{23.85} & \textbf{36.20} & \textbf{94.11} & \textbf{90.63} & \textbf{88.86} & \textbf{92.51} \\
\hline
\end{tabular}
\end{table*}

\begin{figure*}
    \centering
    \includegraphics[width=0.95\linewidth]{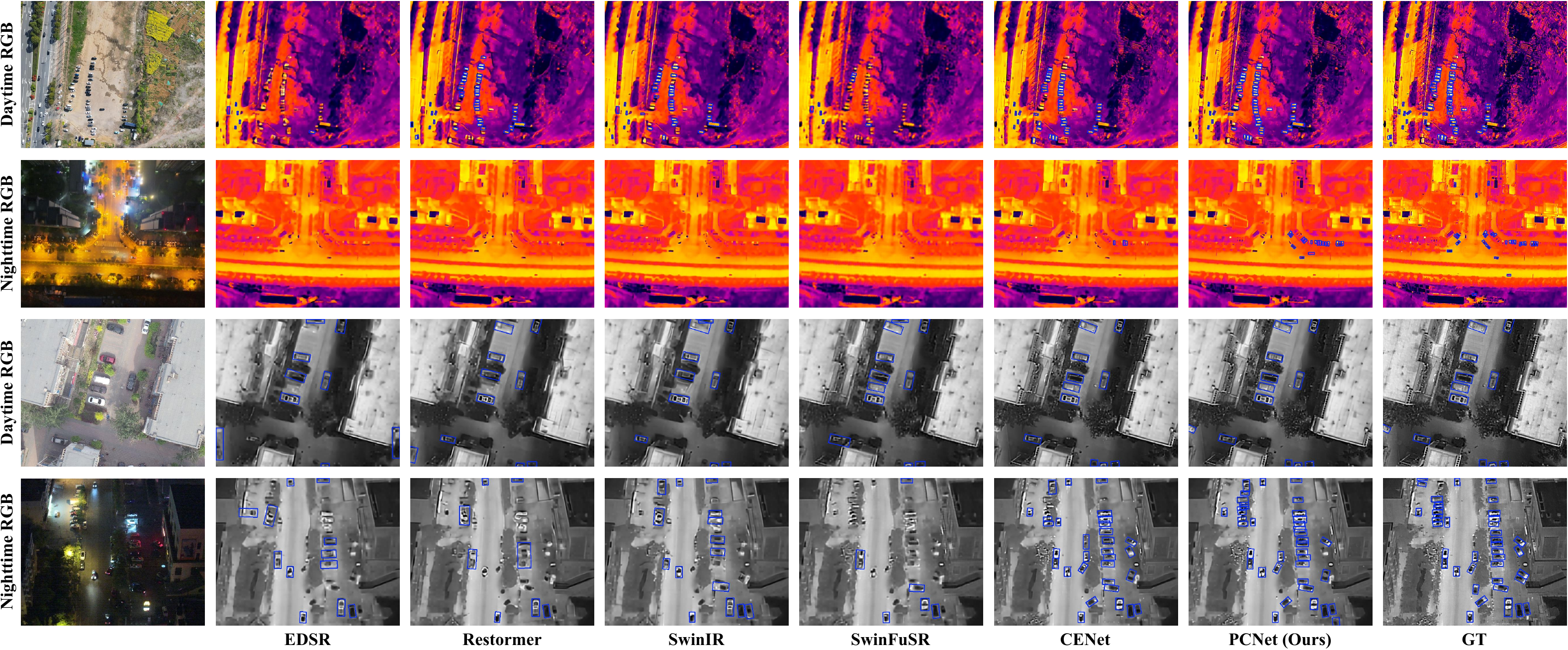}
    \caption{Visual comparison of object detection results after SR using different methods. The first two rows show vehicle detection results on the VGTSR2.0 dataset, while the last two rows present results on the DroneVehicle dataset.}
    \label{detection}
\end{figure*}

Furthermore, the SR results obtained by PCNet exhibit the closest temperature distribution to the ground truth. As shown in Fig.~\ref{temperature}, PCNet demonstrates superior 3D thermal reconstruction with consistent spatial patterns and minimal artifacts compared to competing methods. The temperature and gradient difference visualizations quantitatively validate the effectiveness of PCNet. Specifically, PCNet reduces temperature error by 41\% compared to Bicubic, 11\% compared to EDSR, and 4\% compared to SwinIR. In terms of gradient fidelity, PCNet achieves 34\% reduction compared to Bicubic, 11\% compared to EDSR, and 6\% compared to SwinIR, demonstrating superior thermal field reconstruction capability.

\subsection{Comparison on Downstream Tasks}
\label{Quantitative2}
To validate the practicality of PCNet, we conduct experiments on downstream tasks, specifically focusing on semantic segmentation and object detection using SR results. For the segmentation task, we employ the Segment Anything Model (SAM) ~\cite{Kirillov2023segment} to perform semantic segmentation on the super-resolved results from all competing methods. The segmentation performance is evaluated using four widely adopted metrics: IoU, Dice, ACC, and ASSD. For object detection, we utilize a pre-trained YOLOv8s model~\cite{Jocher2023yolov8} with Oriented Bounding Boxes (OBB) trained on comprehensive vehicle detection datasets, evaluating performance using mAP$_{50}$, mAP, Recall, and F1 metrics.

Table~\ref{tab:seg_vgtsr} and~\ref{tab:seg_drone} present the quantitative comparison of segmentation results on the VGTSR2.0 and DroneVehicle datasets, respectively. On the VGTSR2.0 dataset, our proposed PCNet achieves superior performance across all evaluation metrics. Specifically, PCNet attains an IoU of 69.54\%, representing an improvement of 0.25 percentage points over the second-best performer CENet(69.29\%). Similarly, PCNet demonstrates remarkable performance in Dice of 80.02\%, ACC of 81.50\%, and ASSD of 12.21, consistently outperforming all competing methods. On the DroneVehicle dataset, PCNet achieves the best results with an IoU of 65.21\% and demonstrates strong performance across other metrics, validating its effectiveness in complex vehicular scenarios.

As shown in Table~\ref{tab:detection}, our proposed PCNet significantly outperforms all competing methods across all evaluation metrics in the object detection task. Notably, PCNet achieves 49.60\% mAP$_{50}$, representing a substantial improvement of 4.94 percentage points over the second-best performer CENet (44.66\%). For mAP, PCNet demonstrates exceptional performance of 40.14\%, surpassing CENet (35.49\%) by 4.65 percentage points. On the DroneVehicle dataset, PCNet maintains its superior performance with 94.11\% mAP$_{50}$ and 90.63\% mAP, significantly outperforming all competing methods and demonstrating robust detection capabilities across diverse environmental conditions.

To further validate the effectiveness of our approach, we conduct qualitative comparisons on downstream tasks. Fig.~\ref{segmentation} illustrates the visual comparison of segmentation results after super-resolution reconstruction using different methods. The first two rows represent segmentation results from the VGTSR2.0 dataset, while the latter two rows show results from the DroneVehicle dataset. Our proposed PCNet demonstrates superior structural preservation and clearer object boundaries compared to other approaches. The yellow highlighted regions particularly showcase areas where PCNet achieves more accurate detail reconstruction, approaching the ground truth quality more closely than competing methods. Fig.~\ref{detection} presents the visual comparison of object detection results after super-resolution using different methods. The comparison includes challenging scenarios across both datasets, featuring various lighting conditions and environmental complexities. The proposed PCNet consistently demonstrates superior vehicle detection performance, particularly in preserving fine details and maintaining accurate object localization. The results clearly indicate that PCNet-enhanced images enable more reliable object detection, with improved precision in identifying vehicle boundaries and reducing false negatives compared to other super-resolution methods.

\begin{figure*}
    \centering
    \includegraphics[width=0.9\linewidth]{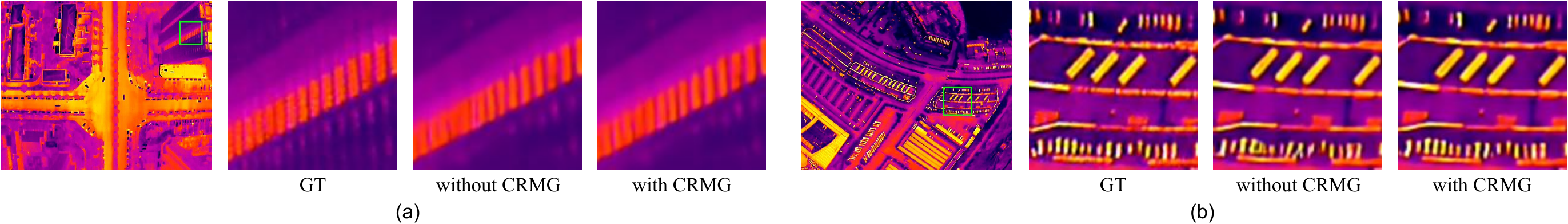}
    \caption{Visual comparison of $\times$4 SR results with and without CRME on the VGTSR2.0 test set. (a) Thermal Images of Skyscraper Windows. (b) Thermal image of dense vehicle traffic.}
    \label{ab_crme}
\end{figure*}
\begin{figure*}
    \centering
    \includegraphics[width=0.9\linewidth]{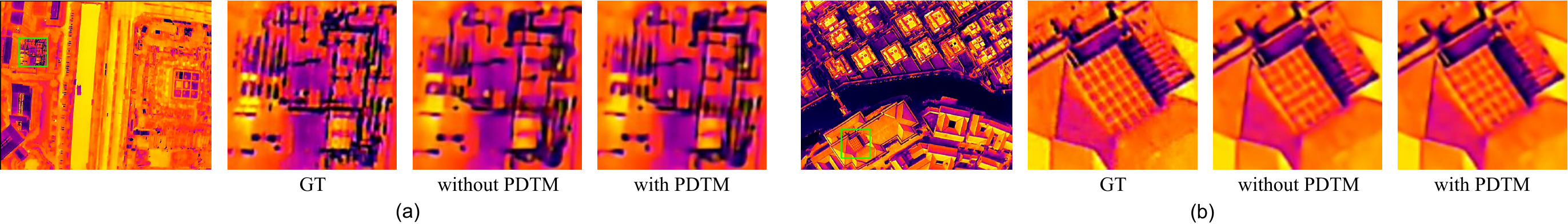}
    \caption{Visual comparison of $\times$4 SR results with and without PDTM on the VGTSR2.0 test set. (a) Thermal image of the ground in front of the high-rise building. (b) Thermal image of the rooftop solar panels.}
    \label{ab_pdtm}
\end{figure*}

\begin{table}
\renewcommand{\arraystretch}{1.5}
\scriptsize
\begin{center}
\caption{Ablation Study of the HTL, CRME, PDTM and TCLoss. Bold font indicates the best performance (SR Scale Factor 4).}
\begin{tabular}{ccccccccccc} 
\toprule
HTL & CRME & PDTM & TCLoss  & PSNR$\uparrow$  & SSIM$\uparrow$  & NIQE$\downarrow$ 
\\ \midrule
$\checkmark$& & &        &31.35  &0.9112 &6.0636  \\ 
$\checkmark$&$\checkmark$ & &       &31.59 &0.9167 &5.9281 \\
$\checkmark$& &$\checkmark$ &       &31.52 &0.9157 &5.9112\\ 
$\checkmark$& & &$\checkmark$       &31.53 &0.9153 &5.9408 \\ \midrule
$\checkmark$&$\checkmark$ &$\checkmark$ &       &31.63 &0.9177 &5.9219 \\
$\checkmark$&$\checkmark$ & &$\checkmark$       &31.62 &0.9178 &\textbf{5.8635} \\
$\checkmark$& &$\checkmark$ &$\checkmark$       &31.54 &0.9161 &5.9592 \\ \midrule
$\checkmark$ &$\checkmark$ &$\checkmark$ &$\checkmark$       &\textbf{31.76} &\textbf{0.9200} &5.9028 \\ \bottomrule
\end{tabular}
\label{ablation1}
\end{center}
\end{table}

\subsection{Ablation Experiments}
\label{Ablations}

\begin{figure*}
    \centering
    \includegraphics[width=0.9\linewidth]{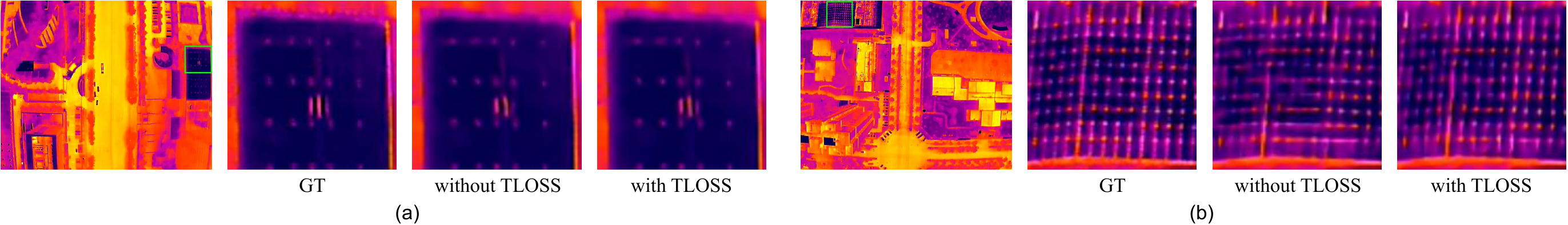}
    \caption{Visual comparison of $\times$4 SR results with and without TCLoss on the VGTSR2.0 test set. (a) Thermal image of the flat roof. (b) Thermal Image of the roof ventilation grille.}
    \label{ab_TCLoss}
\end{figure*}

\vspace{0.5em}
\noindent\textbf{Effectiveness of Cross-Resolution Mutual Enhancement Module.}
To evaluate the effectiveness of CRME in preserving high-frequency optical priors while enabling cross-resolution feature interaction, we conduct comparative experiments on $\times$4 SR using the VGTSR2.0 dataset with the BI degradation model and performance is assessed using metrics such as PSNR, SSIM, and NIQE. As shown in Table~\ref{ablation1}, incorporating the CRME module results in a PSNR improvement of 0.24 dB compared to the baseline using HTL, an SSIM increase of 0.0055, and a NIQE reduction of 0.1355, which significantly demonstrates the effectiveness of CRME in cross-resolution guided thermal image SR. Furthermore, the joint ablation with PDTM and TCLoss also demonstrates that employing HR optical features to guide LR thermal image SR under physical constraints further enhances the quality of generated thermal images. Specifically, the joint ablation with PDTM results in a 0.28 increase in PSNR, a 0.0065 improvement in SSIM, and a 0.1417 reduction in NIQE. Meanwhile, the joint ablation with TCLoss resulted in a PSNR improvement of 0.27, an SSIM increase of 0.0066, and a significant reduction in NIQE of 0.2001.
As illustrated in Fig.~\ref{ab_crme}, the introduction of CRME demonstrates exceptional detail reconstruction capabilities, such as more orderly window distributions in skyscrapers and more accurate vehicle structure reconstruction.

\vspace{0.5em}
\noindent\textbf{Effectiveness of Physics-Driven Thermal Conduction Module.} 
As shown in Table~\ref{ablation1}, incorporating the PDTM module to constrain optical guidance results in a PSNR improvement of 0.17 dB compared to the baseline using HTL, an SSIM increase of 0.0045, and a NIQE reduction of 0.1524. As illustrated in Fig.~\ref{ab_pdtm}, the inclusion of PDTM leads to reconstructed temperature distributions that exhibit better physical consistency, with better-preserved structural details. This confirms that physics-based constraints effectively suppress artifacts typically induced by unconstrained guidance.

\vspace{0.5em}
\noindent\textbf{Effectiveness of Temperature Consistency Loss.} 
Table~\ref{ablation1} also reports the quantitative contribution of TCLoss. Integrating TCLoss into the baseline model improves PSNR, SSIM, and NIQE by 0.18 dB, 0.0041, and 0.1228, respectively, validating its constraint efficacy. Notably, TCLoss demonstrates strong synergy with PDTM and CRME, and their joint application boosts PSNR to 31.76 dB and SSIM to 0.9200, while reducing NIQE to 5.9028. Visual comparisons in Fig.~\ref{ab_TCLoss} corroborate these findings. TCLoss effectively corrects local temperature deviations, particularly in complex structures such as roof surfaces and ventilation grilles. By suppressing artifacts and enhancing texture consistency, it ensures the reliability of the generated images.

\begin{figure*}    
\centering    
    \includegraphics[width=0.85\linewidth]{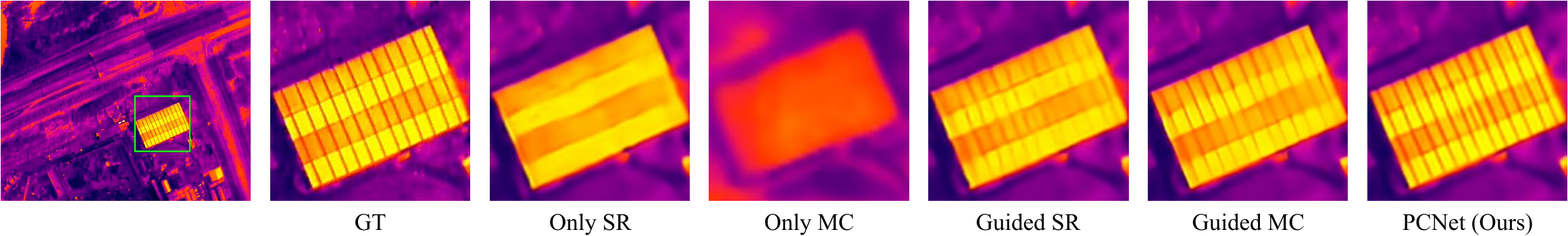}    
\caption{Visual comparison of $\times$8 SR results using different tasks on the VGTSR2.0 test set. The collaborative training framework we propose for MC and SR tasks achieves the best visual results.}
\label{ab_mcsr}
\end{figure*}

\vspace{0.5em}
\noindent\textbf{Effectiveness of Collaborative Enhancement Framework.}
The proposed PCNet leverages the CRME module to facilitate the joint training of MC and SR tasks. To verify the efficacy of this collaborative framework, we conduct an ablation study comparing five training configurations: Only SR, Only MC, Guided SR, Guided MC, and our complete collaborative model. Table~\ref{ablation2} presents the quantitative results on the VGTSR2.0 test set at $\times$8 SR using BI degradation. As observed, training solely on the SR task limits performance, yielding a PSNR of 25.58~dB and SSIM of 0.7636, while the Only MC setting results in significantly degraded reconstruction quality. Incorporating guidance mechanisms into single-task training improves performance, with Guided SR and Guided MC achieving PSNR of 26.16~dB and 26.05~dB, respectively. However, these variants still fall short of the proposed collaborative framework. Our full method achieves the best quantitative metrics with a PSNR of 26.54~dB, SSIM of 0.8003, and NIQE of 6.7165. These results confirm that jointly optimizing MC and SR provides substantial and consistent gains over single-task or simply guided alternatives.
Visual comparisons in Fig.~\ref{ab_mcsr} further corroborate these findings. Reconstructions from Only SR or Only MC exhibit noticeable blurring and structural artifacts. Although single-guided methods alleviate some artifacts, they fail to sufficiently recover fine thermal details. In contrast, collaborative training restores sharper textures and coherent local features, generating images closest to the ground truth.

\begin{table}
\renewcommand{\arraystretch}{1.5}
\scriptsize
\caption{Ablation Study on the Effectiveness of Collaborative Learning Frameworks. Bold font indicates the best performance (SR Scale Factor 8).}
\begin{tabular}{lcccc} 
\toprule
Method &Scale& PSNR$\uparrow$  & SSIM$\uparrow$  &NIQE$\downarrow$ 
\\ \hline
Only SR &8     &25.58  &0.7636 &7.4298  \\
Only MC &8     &12.97 &0.4242 &9.9216 \\
Guided SR &8     &26.16 &0.7857 &6.7939  \\
Guided MC &8     &26.05 &0.7832 &6.8930  \\ \midrule
Ours &8     &\textbf{26.54} &\textbf{0.8003} &\textbf{6.7165} \\ \bottomrule
\end{tabular}
\label{ablation2}
\end{table}

\section{Conclusion}
In this paper, we have addressed the limitations of existing optics-guided thermal image super-resolution methods for UAVs by proposing the Physics-Constrained Cross-Resolution Enhancement Network (PCNet). The proposed framework comprises three core components: the Cross-Resolution Mutual Enhancement Module (CRME), which enables effective bidirectional feature interaction across different resolutions while preserving high-frequency optical priors; the Physics-Driven Thermal Conduction Module (PDTM), which constrains optical images to guide thermal image generation according to thermal physics principles; and the Temperature Consistency Loss, which ensures generated images align with ground truth in temperature distribution and thermal diffusion characteristics. To the best of our knowledge, this is the first physics-constrained cross-resolution network for optics-guided thermal super-resolution. This architecture achieves robust interaction between optical and thermal modalities while preserving the physical consistency and interpretability of generated thermal images. Extensive quantitative and qualitative experiments demonstrate that PCNet not only significantly outperforms state-of-the-art SISR and GISR methods on the VGTSR2.0 and DroneVehicle datasets but also exhibits robust performance in downstream tasks. Future research will explore non-registration-guided super-resolution methods inspired by distinct physical properties.

\section*{Declaration of Competing Interest}
The authors declare that they have no known competing financial interests or personal relationships that could have appeared to influence the work reported in this paper.

\section*{Acknowledgement}
This work was supported in part by the National Natural Science Foundation of China (No. 62306005, 62006002, and 62076003), in part by the Joint Funds of the National Natural Science Foundation of China (No. U20B2068), in part by the Natural Science Foundation of Anhui Province (No. 2208085J18 and 2208085QF192), and in part by the Natural Science Foundation of Anhui Higher Education Institution (No. 2022AH040014).

\bibliographystyle{unsrt}

\bibliography{ref}
\end{sloppypar}
\end{document}